\documentclass{amsart}
\usepackage{amsmath}%
\usepackage{amsfonts}%
\usepackage{amssymb}%
\usepackage{graphicx}
\usepackage{algpseudocode}
\usepackage{algorithmicx}
\usepackage{algorithm}
\usepackage{color}
\usepackage{caption}
\usepackage[normalem]{ulem}

\newcommand{\norm}[1]{\left\Vert#1\right\Vert}
\newcommand{\abs}[1]{\left\vert#1\right\vert}
\newcommand{\set}[1]{\left\{#1\right\}}
\newcommand{\Real}{\mathbb R}
\newcommand{\Prob}{\mathbb P}
\newcommand{\vareps}{\varepsilon}

\newcommand{\bareps}{\bar{\epsilon}}

\newcommand{\vecx}{\mathbf x}
\newcommand{\vecb}{\mathbf b}
\newcommand{\vecs}{\mathbf s}
\newcommand{\vecy}{\mathbf y}
\newcommand{\vecz}{\mathbf z}
\newcommand{\veca}{\mathbf a}

\newcommand{\veco}{\mathbf 0}

\newcommand{\matA}{\mathbf A}
\newcommand{\matD}{\mathbf D}
\newcommand{\matE}{\mathbf E}
\newcommand{\matG}{\mathbf G}
\newcommand{\matX}{\mathbf X}
\newcommand{\matY}{\mathbf Y}
\newcommand{\matV}{\mathbf V}
\newcommand{\matU}{\mathbf U}
\newcommand{\matI}{\mathbf I}
\newcommand{\vecnoise}{\mathbf \vareps}
\newcommand{\vecxi}{\mathbf \xi}
\newcommand{\vecxij}{\mathbf \xi_j}
\newcommand{\vecphi}{\mathbf \phi}
\newcommand{\vecphij}{\mathbf {\phi}_{i,j}}
\newcommand{\veczeta}{\mathbf \zeta}
\newcommand{\gideriv}{g^{\prime}_i}
\newcommand{\gjderiv}{g^{\prime}_j}
\newcommand{\gijderiv}{g^{\prime\prime}_{ij}}
\newcommand{\gilderiv}{g^{\prime\prime}_{il}}
\newcommand{\gjlderiv}{g^{\prime\prime}_{jl}}
\newcommand{\giloderiv}{g^{\prime\prime}_{il_1}}

\newcommand{\gjltderiv}{g^{\prime\prime}_{jl_2}}
\newcommand{\setX}{\mathcal X}
\newcommand{\setPhi}{\mathbf \Phi}

\newtheorem{theorem}{Theorem}
\theoremstyle{plain}

\newtheorem{corollary}{Corollary}

\newtheorem{lemma}{Lemma}

\newtheorem{prop}{Proposition}
\newtheorem{remark}{Remark}

\numberwithin{equation}{section} 

\title[Learning non-parametric basis independent models]{Learning non-parametric basis independent models from point queries via low-rank methods}

\author{Hemant Tyagi}
\curraddr[H. Tyagi]{ETH Z\"{u}rich,}
\email[H. Tyagi]{htyagi@inf.ethz.ch}%
\author{Volkan Cevher}
\curraddr[V. Cevher]{Laboratory for Information and Inference Systems (LIONS),\newline%
\indent Ecole Polytechnique F\'{e}d\'{e}rale de Lausanne,}%
\email[V. Cevher]{volkan.cevher@epfl.ch}%

\thanks{An extended abstract of this paper appeared in the $26^{\text{th}}$ Annual Conference on Neural Information Processing Systems (NIPS), December 2012. 
The present draft is an expanded version with a more rigorous analysis and consists of proofs of all the results.}


\begin{document}

\maketitle
%
\begin{abstract}
We consider the problem of learning \textit{multi-ridge} functions of the form
$f(\vecx) = g(\matA\vecx)$ from point evaluations of $f$. We assume that the
function $f$ is defined on an $\ell_2$-ball in $\Real^d$, $g$ is twice
continuously differentiable almost everywhere, and $\matA \in \mathbb{R}^{k
\times d}$ is a rank $k$ matrix, where $k \ll d$. We propose a randomized,
polynomial-complexity sampling scheme for estimating such functions. Our
theoretical developments leverage recent techniques from low rank matrix
recovery, which enables us to derive a polynomial time estimator of the function
$f$ along with uniform approximation guarantees. We prove that our scheme can
also be applied for learning functions of the form: $f(\vecx) = \sum_{i=1}^{k}
g_i(\veca_i^T\vecx)$, provided $f$ satisfies certain smoothness conditions in a
neighborhood around the origin. We also characterize the noise robustness of the
scheme. Finally, we present numerical examples to illustrate the theoretical
bounds in action.
\end{abstract}

recovery, randomized sampling, oracle-based learning

%
\section{Introduction}\label{sec: intro}
\noindent Many important scientific and engineering problems revolve around models defined as multivariate continuous functions of $d$ variables, where $d$ is typically large. Examples include but are not limited to neural networks that are commonly used in pattern classification from data \cite{Bishop1996}, path integrals with respect to Weiner measure that arise in the parameter estimation of stochastic processes~\cite{Muller2008}, and smooth multivariate objective functions in optimization problems in machine learning and signal processing. As having an explicit form of a multivariate continuous function $f$ alleviates analysis and computation in many applications, a great deal of research now focuses on learning such functions from their point values \cite{Cohen2010,Fornasier2010,Raskutti2010}.

Unfortunately, even approximating multivariate continuous functions defined over classical unweighted spaces is in general intractable. This notion of intractability is precisely characterized by the \textit{information complexity} of learning, which is defined as the minimum number of information extraction operations $n(e,d)$ that an algorithm performs to estimate a multivariate function within a uniform approximation error $e$~\cite{Traub1988}. If $n(e,d)$ depends exponentially on either $e^{-1}$ or $d$, then the problem is called intractable. Polynomial tractability, on the other hand, specifically refers to the case when $n(e,d)$  depends polynomially on both $d$ and $e^{-1}$. In the function learning setting, it is well known that the optimal order of the error of approximation for functions belonging to $\mathcal{C}^r [0,1]^d$ is exponential: i.e., $n(e,d) = \Omega((1/e)^{d/r}$) for $e \in (0,1)$ (see \cite{Traub1988} for example). As another example, \cite{Novak2009} recently proved that the $L_{\infty}$ approximation of $\mathcal{C}^{\infty}$ functions defined on $[0,1]^d$ is an intractable problem: i.e., $n(e,d) = \Omega(2^{\lfloor d/2 \rfloor})$ for $e \in (0,1)$. Therefore, further assumptions on the multivariate functions beyond smoothness are needed for the tractability of successful learning  \cite{hardle1990applied,Fornasier2010,Cohen2010,Traub1988}.

Fortunately, many multivariate functions that arise in practice possess much more structure than an arbitrary $d$-variate continuous function. To this end, our work focuses on approximating a particular class of low dimensional functions known as \textit{multi-ridge functions} with point queries. A multi-ridge function is a multivariate function $f: \ \mathbb{R}^d \rightarrow \mathbb{R}$ defined using a $k \times d$, full rank matrix $\matA$ as follows:
\begin{equation}
f(\vecx) \ = \ g(\matA\vecx) \label{eq:ridge_functions},
\end{equation}
where $g$ belongs to a restricted function class. Ridge functions are studied in Statistics under the name of ``projection pursuit regression''~\cite{Friedman81,Donoho89,Huber85}. The namesake was first introduced for the case $k=1$ in 1975 by Logan and Shepp ~\cite{Logan75}, in connection with the mathematics of computer tomography.  Approximation theoretical questions regarding ridge functions have been studied in connection with the modeling of neural networks~\cite{Pinkus99,Candes99a}, and also in ridgelets~\cite{Candes99b,Candes03}. A special case of \eqref{eq:ridge_functions} where $f$ decomposes as:
\begin{equation}
f(\vecx) \ = \ \sum_{i=1}^{k} g_i(\veca_i^T \vecx), \label{eq:mult_index_models}
\end{equation}
has several important applications in machine learning applications and are known as \textit{multi-index} models in statistics and econometrics \cite{xia2002adaptive,xia2008multiple,li1991sliced,hall1993almost}.

\paragraph{\bf Previous work.} The recent literature can be split into two distinct camps with one taking an approximation theoretic view and the other pursuing a regression perspective.

In the approximation theoretic camp, the data is obtained with a sampling strategy tailored towards the structure of the underlying function $f$. \cite{Cohen2010} propose a greedy algorithm for estimating functions of the form $f(\vecx) \ = \ g(\veca^T\vecx)$, where $g:[0,1] \rightarrow \Real$ is a $\mathcal{C}^s$ function for $s \ge 1$. To establish tractable learning guarantees on $f$, the authors assume that $\veca$ is stochastic, that is,  $\veca \succeq 0$ and $\mathbf{1}^T\veca = 1$. They also assume $\veca$ to be compressible, i.e., $\veca$ lives in a weak $\ell_q$-ball, and hence, can be well-approximated by a sparse set of its coefficients. In ~\cite{Fornasier2010}, the authors generalize the model of Cohen et al.\ to the matrix case \eqref{eq:ridge_functions} by assuming that each row of $\matA$ is compressible without any sign restrictions and that $g$ is in $\mathcal{C}^s$ for $s\ge 2$.

In the regression camp, the data is drawn independent and identically distributed (iid) from some unknown distribution. \cite{Raskutti2010} leverage convex programming based on $M$-estimators, and study the sparse additive model, $f(\vecx) = \sum_{j \in S} g_j(x_j)$ ($\abs{S}=k\ll d$), introduced by \cite{lin2006component}. In this setting, \cite{Raskutti2010} remove the smoothness assumptions on the function atoms $g_j$, and treat the case where $g_j$'s lie in a reproducible Hilbert Kernel space. Moreover, \cite{Raskutti2010} provide algorithm independent minimax approximation rates. For more examples in the regression camp, we refer the reader to \cite{ravikumar2009sparse,koltchinskii2010sparsity,meier2009high,lin2006component}.

\paragraph{\bf Our contributions.} These works rigorously illustrate that it is highly advantageous to identify additional structures in the multivariate function for the tractability of learning. In this setting, our work belongs to the approximation theoretic camp and makes the following three contributions.

First, we generalize the approximation results of \cite{Fornasier2010} to the class of $\mathcal{C}^2$ functions with arbitrary number of linear parameters $k$ \textit{without} the compressibility assumption on the rows of $\matA$. To achieve this generalization, we leverage recent advances in the analysis of low-rank matrix recovery algorithms. As a result, we propose a stable, polynomial time algorithmic framework with a tractable sampling scheme, endowed with uniform approximation guarantees on $f$.

Second, we prove tractability of our framework for a wider function class - a key addition to the existing results which are limited to radial functions \cite{Fornasier2010}. To acheive this we place second order conditions on $f$ which are made clear in Proposition \ref{prop:alpha_tract_gen}. As a side result, we are able to handle the important case of multi-index models \eqref{eq:mult_index_models}. For instance, summation of $k$-kernel ridge functions (Epanechnikov, Gaussian, Cosine, etc.) functions are readily handled. This result also lifts the structure of sparse additive model from the regression camp to a basis free setting, but in turn restricts the functional atoms to be almost everywhere $\mathcal{C}^2$. 

Third, we empirically illustrate the tightness of our sample complexity bounds on a variety of important function examples, such as logistic, quadratic forms, and summation of Gaussians. We also analytically show how additive white noise in the function queries impacts the sample complexity of our low-rank based approach.

%
\paragraph{\bf Notation.} We denote the $\ell_2$-ball with radius $r>0$ in $\Real^{d}$  as $B_{\Real^{d}}(r)$, and employ the shorthand $B_{\Real^{d}}$ when $r=1$.  We use $\mu_{\mathbb{S}^{d-1}}$ for the uniform measure on the $d$-dimensional unit sphere $\mathbb{S}^{d-1}$. For $\vecx,\vecy \in \Real^{d}$, we let $\langle\vecx,\vecy\rangle \ = \ \vecx^T\vecy$ denote the inner product. We use $\ll\matX,\matY\gg \ = \ \text{Tr}(\matX^T\matY)$ as the standard matrix inner product where $\text{Tr}(\cdot)$ is the matrix trace. $\norm{\matX}_{*}$ denotes the nuclear norm, $\norm{\matX}_{F}$ denotes the Frobenius norm, and $\norm{\matX}$ denotes the operator norm of $\matX$. For any $\vecx \ \in \ \Real^n$ we denote its $\ell_p$ norm by $\norm{\vecx}_{\ell_p^n}$. For a given linear operator $\Phi: \Real^{n_1 \times n_2} \rightarrow \Real^{m}$, we use $[\Phi(\matX)]_i = \ll\Phi_i,\matX\gg$ with $\Phi_i \in \Real^{n_1 \times n_2}$, and denote $\Phi^{*}: \Real^{m} \rightarrow \Real^{n_1 \times n_2}$ as the adjoint operator. 
%
\vspace{-3mm}
\section{Setup and Assumptions} \label{sec: problem_setup}

\paragraph{\bf Problem statement.} Broadly speaking, we are interested in deriving approximations for functions $f: B_{\Real^{d}}(1+\bareps) \rightarrow \Real$ of the form
$f(\vecx) = g(\matA\vecx)$, where $\matA = [\veca_1,\dots,\veca_k]^T$ is an arbitrary rank $k$ matrix of dimensions $k \times d$. We restrict ourselves to the \emph{oracle} setting where we can only extract information about $f$ through its---possibly noisy---point evaluations.

\paragraph{\bf Assumptions.} We first assume $\matA\matA^T = \matI_{k}$, where $\matI$ is the $k\times k$ identity matrix. If this is not the case, we can express $\matA$ through its singular value decomposition (SVD) as $\matA = \matU\Sigma\matV^T$ to obtain an equivalent representation: $f(\vecx) \ = \ g(\matU\Sigma\matV^T\vecx) \ = \ \bar{g}(\matV^T\vecx)$, where
$\bar{g}(\vecy) =  g(\matU\Sigma\vecy)$ and $\vecy \in B_{\Real^k}(1+\bareps)$. It is straightforward to verify how our assumptions on $g$ transfers on $\bar{g}$ (cf., \cite{Fornasier2010}). While we discuss approximation results on $A$ below, the readers should keep in mind that our final guarantees only apply to the function $f$ and not necessarily for $\matA$ and $g$ individually.

We assume $g$ to be a $\mathcal{C}^2$ function. 
By our set up, $g$ also lives over a compact set, hence all its partial derivatives till the order of two are bounded as a result of the Stone-Weierstrass theorem:
\begin{equation*}
\text{sup}_{\abs{\beta} \leq 2} \norm{D^{\beta}g}_{\infty} \leq C_2;~~~ D^{\beta} g \ = \ \frac{\partial^{\abs{\beta}}}{\partial y_1^{\beta_1} \dots \partial y_k^{\beta_k}} \ ; \quad \abs{\beta} = \beta_1 + \dots + \beta_k
\end{equation*}
for some constant $C_2  > 0$. We also assume that an enlargement of the unit ball $B_{\Real^{d}}$ on the domain of the function $f$ for a sufficiently small $\bar{\epsilon} > 0$ is allowed. This is not a restriction, but is a consequence of our analysis as we work with directional derivatives of $f$ at points on the unit sphere $\mathbb{S}^{d-1}$.

\paragraph{\bf Our Ansatz.} We verify the tractability of our sampling approach by checking whether or not the following Hessian matrix $H$ is well-conditioned \'a la \cite{Fornasier2010}:
\begin{equation} \label{eq:mat_cond_param}
H^{f} := \int_{\mathbb{S}^{d-1}}\nabla f(\vecx) \nabla f(\vecx)^{T} d\mu_{\mathbb{S}^{d-1}}(\vecx).
\end{equation}
That is, for singular values of $H^{f}$, we have $\sigma_1(H^{f}) \geq \sigma_2(H^{f})\geq \dots \geq \sigma_k(H^{f}) \geq \alpha > 0$ for some $\alpha$. We theoretically characterize the scaling of $\alpha$ in Section \ref{sec:tract_samp_complexity} for interesting classes of functions.

\section{Oracle-based Low-Rank Learning of Multi-Ridge Functions}
In this section, we first identify a first-order relationship in our learning
problem that ties the function values at the point queries as an affine
observation of a low-rank matrix, whose column space is equal to $A^T$. We then
exploit this observation to motivate a class of polynomial time algorithms for
approximate recovery of $A$. To establish algorithmic guarantees, we focus on a
randomized sampling scheme that provides a bi-Lipschitz embedding of low rank
matrices. We then provide an outline of our learning scheme, which we
theoretically analyze in Section \ref{sec: analysis}. 
\subsection{{Observation and oracle models}}
Our learning approach relies on a specific interaction of two sets: sampling
centers and an associated set of directions for each center. Let us first denote
the set of sampling centers as follows:
\begin{equation} \label{eq:uniform_set}
\setX = \{\vecxi_{j} \in \mathbb{S}^{d-1}; j=1,\dots,m_{\setX}\}.
\end{equation}
Along with each $\vecxi_{j}\in \setX$, we define a directions matrix $\Phi_j =
\left[{\vecphi}_{1,j} \vert \ldots \vert {\vecphi}_{m_{\Phi},j}\right]^T$, where
$\vecphi \in B_{\Real^{d}}(r)$ for some $r>0$, which we specify in Section
\ref{sec: sampling}.

We now begin with a simple first order approximation of the function $f$ as
follows
\begin{equation}\label{eq: first order relation}
  f(\vecx+\epsilon\phi) = f(\vecx) + \epsilon \left\langle \phi, \nabla f(\vecx)
\right\rangle + \epsilon E(\vecx,\epsilon,\phi),
\end{equation}
where $\epsilon\ll 1$, and $\epsilon E(\vecx,\epsilon,\phi)$ is the
approximation error. Substituting the ridge function form
\eqref{eq:ridge_functions} into \eqref{eq: first order relation}, we then
stumble upon a perturbed observation model ($\nabla g(\cdot)$ is a $k\times 1$
vector) below
\begin{equation}\label{eq: first order ridge relation}
  \left\langle \phi, A^T\nabla g(A\vecx)\right\rangle =
\frac{1}{\epsilon}\left(f(\vecx+\epsilon\phi)- f(\vecx)\right) - 
E(\vecx,\epsilon,\phi).
\end{equation}

Without loss of generality, we denote the evaluation of $f(\vecx+\epsilon\phi)-
f(\vecx)$ as a call to the oracle. When the oracle is flawless, then the error
$E(\vecx,\epsilon,\phi)$ is characterized via Taylor's expansion:
\begin{equation}
  E(\vecx,\epsilon,\phi)= \vecnoise :=\frac{\epsilon}{2} \phi^T
\nabla^2f(\veczeta(\vecx,\phi))\phi,
\end{equation}
where $\veczeta(\vecxi,\vecphi)\in [\vecxi,\vecx+\epsilon\phi]  \in
B_{\Real^{d}}(1+\epsilon r)$.
In general, one can envision a noisy oracle providing imprecise  function
values. To address a broad set of cases, we modify the perturbation model as
\begin{equation}\label{eq: noisy oracle}
  E(\vecx,\epsilon,\phi)= \vecnoise+ \epsilon^{-1}\vecz + \vecs(\pi),
\end{equation}
where $\vecz = {\mathcal N}(0,\sigma_z^2)$ is an iid, zero mean Gaussian noise
with a variance parameter $\sigma^2_z$, and $\vecs$ is an unbounded
\emph{sparse} noise that either destroys the information in an oracle call with
probability $\pi\ll 1$, or leaves it untouched with probability $1-\pi$. Section
\ref{subsec:imp_meas_noise} further addresses the noise issues.

\subsection{{Low-rank matrix recovery of $A$}} \label{subsec:lowrank_recov}
We now leverage \eqref{eq: first order ridge relation} as a scaffold to derive
our low-rank learning approach. We first  introduce a rank-$k$ matrix $\matX :=
\matA^T\matG$ with $\matG := [\nabla g(\matA\vecxi_1) \vert \nabla
g(\matA\vecxi_2) \vert \cdots \vert \nabla g(\matA\vecxi_{m_{\setX}})]_{k \times
m_{\setX}}$.
Based on \eqref{eq: first order ridge relation}, we then derive the following
linear system of equations via the linear operator  $\Phi: \Real^{d \times
m_{\setX}} \rightarrow \Real^{m_{\Phi}}$
\begin{equation}\label{factorization_equation_gen}
\vecy = \Phi(\matX) + E(\setX,\epsilon,\setPhi),
\end{equation}
where we refer to $\vecy\in \Real^{m_{\Phi}}$ as the (perturbed) measurements of
$\matX$.

The formulation \eqref{factorization_equation_gen} is known as the low-rank
matrix recovery problem since the rank of the matrix $\matX$ is $k\ll d$. In
Appendix \ref{sec: ap: lowrank}, we explain three distinct low-rank recovery
problem settings relevant to our problem, called affine rank minimization (ARM),
matrix completion (MC), and robust principal component analysis (RPCA). Among
these low-rank formulations, we focus on a randomized sampling scheme for the
ARM problem using the matrix Dantzig selector for our derivations below. We
leave the theoretical characterization the subset selection schemes for future.

\subsection{Low-rank matrix sampling}\label{sec: sampling}
It turns out that stable recovery of $\matX$ from
\eqref{factorization_equation_gen} is provable from number of measurements
commensurate with the degrees of freedom in $\matX$ (i.e., $m_{\Phi}={\mathcal
O}\left(k(d+m_{\setX}-k)\right)$). By stable, we mean that the error of the
estimated matrix in Frobenius norm is bounded by a constant times the Frobenius
norm of the perturbations. Moreover, via the RPCA formulation, it is also
possible to stably recover $\matX$ even when a fraction of its entries are
arbitrarily corrupted. These recovery guarantees of course are predicated upon
the sampling scheme preserving the information in the low-rank matrix.

For concreteness, we require our sampling mechanism in this paper to provide a
bi-Lipschitz embedding of all rank-$r$ matrices $\matX_r$ with overwhelming
probability:
\begin{equation*}
(1-\kappa_r)\norm{\matX_r}_F^2 \leq \norm{\Phi(\matX_r)}^2_{l_2} \leq
(1+\kappa_r)\norm{\matX_r}_F^2,
\end{equation*}
where $\kappa_r$ is known as the the isometry constant \cite{Candes2010}. We say
that $\Phi$ satisfies the $\kappa$-RIP at rank $r$ if $\kappa_r < \kappa$ where
$\kappa\in (0,1)$. For the linear operator $\Phi$ to have $\kappa$-RIP, we form
$\setX$ by sampling points uniformly at random in $\mathbb{S}^{d-1}$ according
to the uniform measure $\mu_{\mathbb{S}^{d-1}}$. We then construct the sampling
directions for $i=1,\dots,m_{\Phi}, \ j=1,\dots,m_{\setX}, \ \text{and} \ l =
1,\dots, d$ as follows
\begin{equation}
\setPhi = \left\{\vecphij \in B_{\Real^{d}}\left(\sqrt{d/m_{\Phi}}\right):
[\vecphij]_l = \pm \frac{1}{\sqrt{m_{\Phi}}} \text{with probability} \ 1/2
\right\}. \label{eq:direc_deriv_set}
\end{equation}

As $\Phi$ is a Bernoulli random measurement ensemble it follows from standard
concentration inequalities ~\cite{RechtFazel2010,Laurent2000} that for any
rank-$r$ $\matX \in  \Real^{d \times m_{\setX}}$
\begin{equation*}
\Prob(\vert\norm{\Phi(\matX)}_{\ell_2}^2 - \norm{\matX}_F^2\vert >
t\norm{\matX}_F^2) \leq 2e^{-\frac{m_{\Phi}}{2}(t^2/2 -t^3/3)}, \quad  t \in
(0,1).
\end{equation*}
By using a standard covering argument as shown in Theorem 2.3 of
~\cite{Candes2010} it is easily verifiable that $\Phi$ satisfies RIP with
isometry constant $0 < \kappa_r < \kappa < 1$ with probability at least
$1-2e^{-m_{\Phi}q(\kappa) + r(d+m_{\setX}+1)u(\kappa)}$, where $q(\kappa) =
\frac{1}{144}\left(\kappa^2 -\frac{\kappa^3}{9}\right)$ and $u(\kappa) =
\log\left(\frac{36\sqrt{2}}{\kappa}\right)$.

%
\subsection{Our low-rank oracle learning scheme}
We outline the main steps involved in our approximation scheme in Algorithm
\ref{alg:ridge_approx}. Step 1 is related to the sampling tractability of
learning, which we study in Section \ref{sec:tract_samp_complexity}. Step 2
forms the measurements based on the ARM formulation and our sampling scheme.
Step 3 revolves around the ARM recovery, where we employ the matrix Dantzig
selector algorithm for concreteness in our analysis. Step 4 maps the recovered
low-rank matrix to $A$, followed by Step 5 that finally leads to the function
estimate.
%
\begin{algorithm}
\caption{Estimating $f(\vecx) = g(\matA\vecx)$} \label{alg:ridge_approx}
\begin{algorithmic}[1]

\State Choose $m_{\Phi}$ and $m_{\setX}$ (Section
\ref{sec:tract_samp_complexity}) and construct the sets $\setX$ and $\setPhi$
(Section \ref{sec: sampling}).

\State Choose $\epsilon$ (Section \ref{subsec:approx_A}) and construct $\vecy$
using $y_i = \sum_{j=1}^{m_{\setX}}\left[\frac{f(\vecxi_j +
\epsilon\vecphij)-f(\vecxi_j)}{\epsilon} \right]$.

\State Obtain $\widehat{\matX}$ via a stable low-rank recovery algorithm
(Appendix \ref{sec: ap: lowrank}).

\State Compute SVD$(\widehat{\matX}) \ = \ \widehat{\matU} \widehat{\Sigma}
\widehat{\matV}^T$ and set $\widehat{\matA}^T = \widehat{\matU}^{(k)}$,
corresponding to $k$ largest singular values.

\State Obtain $\widehat{f}(\vecx) \ := \ \widehat{g}(\hat{\matA} \vecx)$ via
quasi interpolants where $\widehat{g}(\vecy) \ := \ f(\widehat{\matA}^T\vecy)$.
\end{algorithmic}

\end{algorithm}
Section \ref{sec: analysis} provides an end-to-end analysis of the steps in
Algorithm \ref{alg:ridge_approx}. Here, we further comment on two important
ingredients in our learning scheme: the norm of the perturbations, and the
function estimator in Step 5 of Algorithm \ref{alg:ridge_approx} given an
estimate $\widehat{A}$ of $A$.

\paragraph{\bf Stability.} We provide a stability characterization for the ARM
recovery algorithms in the form of Proposition
~\ref{proposition:noise_bounded_proof_gen} below, which upperbounds the
$\ell_2^{m_{\Phi}}$-norm of the noise $\vecnoise$ for the perfect oracle
setting.
%
\begin{prop} \label{proposition:noise_bounded_proof_gen}
In the factorization equality ~\eqref{factorization_equation_gen}, we have
$\norm{E}_{\ell_2^{m_{\Phi}}}=\norm{\vecnoise}_{\ell_2^{m_{\Phi}}} \ \leq \
\frac{\displaystyle C_2 \epsilon k^2 }{2} \frac{ m_{\setX}d}{\displaystyle
\sqrt{m_{\Phi}}}$.
\end{prop}
\noindent Appendix ~\ref{sec:bounded_noise_proof} has the proof. Note that the
dimension $d$ appears in the bound as we do not make any compressibility
assumption on $\matA$. If the rows of $\matA$ are compressible, that is
$(\sum_{j=1}^{d}\abs{a_{ij}}^q)^{1/q}  \ \leq D_1$ $\forall \ i=1,\dots,k$ for
some $0 < q < 1, \ D_1 > 0$, the bound becomes independent of $d$.

\paragraph{\bf Our function estimator.} Given $\widehat{\matA}$ of $\matA$ in
Step 4, we construct $\widehat{f}(\vecx) \ := \ \widehat{g}(\widehat{\matA}
\vecx)$ as our estimator,  where $\widehat{g}(\vecy) \ := \
f(\widehat{\matA}^T\vecy)$ with $\vecy \in B_{\Real^k}(1+\bar{\epsilon})$. We
uniformly approximate the function $\widehat{g}$ by first sampling it on a
rectangular grid : $h\mathbb{Z}^k \cap
(-(1+\bar{\epsilon}),(1+\bar{\epsilon}))^k$ with uniformly spaced points in each
direction (step size $h$). We then using quasi interpolants to interpolate in
between the points thereby obtaining the approximation $\hat{g}_h$, where the
complexity only depends on $k$. We refer the reader to Chapter 12 of
~\cite{devore1993} regarding the construction of these operators.

It is straightforward to prove that $\norm{\widehat{g} - \widehat{g}_h}_{\infty}
< C h^{2}$, holds true for some constant $C$. By triangle inequality, we then
carry the following approximation guarantee for $\widehat{g}_h$:
\begin{equation*}
\norm{g - \widehat{g}_h}_{\infty} \leq \norm{g - \widehat{g}}_{\infty} +
\norm{\widehat{g} - \widehat{g}_h}_{\infty}.
\end{equation*}
In this loop, the samples of $\widehat{g}$ on the $h$-grid are obtained directly
through point queries of $f$. However, the required number of samples for a
given error depends only on $k$ and not on $d$.

\begin{remark} \label{rmk:eps_bar_issue}
(i) The parameter $\bar{\epsilon} = \epsilon\sqrt{d/m_{\Phi}}$, defining the domain
of the function $f$, is bounded from above. In the course of deriving an
approximation to $f$, we require $\epsilon$ to be at most
$\mathcal{O}\left(\frac{1}{d}\sqrt{\frac{m_{\Phi}\alpha}{m_{\setX}}}\right)$,
(as is stated in Lemma ~\ref{lemma:mata_approximation}), in order to obtain a
non trivial approximation error guarantee. We shall also discover in Section
~\ref{sec:tract_samp_complexity} that $\alpha$ can be at most $\mathcal{O}(1)$
implying $\bar{\epsilon}$ to be typically at most $\mathcal{O}(1/\sqrt{d})$. \newline

\noindent (ii) As opposed to \cite{Fornasier2010} our scheme requires more number of sampling directions.
To see this, observe that there is an underlying $d \times m_{\setX}$ matrix $X = A^T G$ which contains information about the gradients of $f$ 
at the sampled points $m_{\setX}$.
Here $\matG := [\nabla g(\matA\vecxi_1) \vert \nabla g(\matA\vecxi_2) \vert \cdots \vert \nabla g(\matA\vecxi_{m_{\setX}})]_{k \times m_{\setX}}$ and $\matA$ 
is the underlying subspace matrix of size $k \times d$. Now in \cite{Fornasier2010}, the compressibility assumption on the rows of $\matA$ enables the authors to sample each 
column of $X$ individually and then recover it using standard $\ell_1$ minimisation.
Note that each column of $X$ is the linear combination of $k$-vectors each of which is compressible hence the resulting $X$ will have compressible columns.
In particular the same direction vector (generated at random) is used for measuring each column of $X$ implying that for $m_{\Phi}$ measurements of the columns
they need only $m_{\Phi}$ sampling directions.
On the other hand we cannot do this since we make no compressibility assumption on $\matA$.
Hence we resort to taking linear measurements of the complete matrix $X$ and aim to recover this matrix by employing low-rank matrix recovery algorithms.
To obtain one measurement of $X$ we need to generate $m_{\setX}$ number of sampling directions implying that for $m_{\Phi}$ measurements of $X$ we need
$m_{\setX} \times m_{\Phi}$ sampling directions.
\end{remark}
\section{Analysis of Oracle-based Low-Rank Learning}\label{sec: analysis}
In this section, the parameters involved our derivations are the dimension $d$ of $\vecx$, the number of linear parameters $k$, the smoothness constant $C_2$ for the underlying function $g$, and the conditioning parameter $0 < \alpha < kC_2^{2}$ for $H^f$ in ~\eqref{eq:mat_cond_param}. Section \ref{sec:tract_samp_complexity} unifies the results with our tractability claims.

%
\subsection{Low-rank matrix recovery with Dantzig Selector}{\label{subsec:dantzig_selector}}
In order to recover an approximation to the rank $k$ matrix $\matX$, we solve the nuclear norm minimization problem based on the following convex formulation ~\cite{Candes2010}:
\begin{equation} \label{eq:MDS}
\widehat{\matX}_{DS}= \arg\min \norm{M}_{*} \text{s.t.}~\norm{\Phi^{*}\left(y - \Phi(M)\right)}\leq \lambda,
\end{equation}
where the optimal solution is the estimate $\widehat{\matX}_{DS}$. This convex program is referred to as the \textit{matrix Dantzig selector} ~\cite{Candes2010}. While Appendix \ref{sec: ap: lowrank} lists a number of other convex formulations for low rank matrix recovery, we choose the matrix Dantzig selector for concreteness.

As in ~\cite{Candes2010}, we require the true matrix $\matX$ to be feasible in the convex formulation, i.e., one should have $\norm{\Phi^{*}(\vecnoise)} \leq \lambda$. In the case of bounded noise, Lemma ~\ref{lemma:adjoint_noise_bound} helps us choose this parameter whose proof is in Appendix \ref{sec:specnorm_adj_noise_proof}.
%
\begin{lemma} \label{lemma:adjoint_noise_bound}
Given $\vecnoise$ with a bounded $\ell_2^{m_{\Phi}}$ norm, it holds that $\norm{\Phi^{*}(\vecnoise)} \leq \frac{\displaystyle C_2 \epsilon d m_{\setX} k^2}{\displaystyle 2\sqrt{m_{\Phi}}}(1+\kappa_1)^{1/2},$ with
probability at least $1-2e^{-m_{\Phi}q(\kappa_1)+(d+m_{\setX}+1)u(\kappa_1)}.$
\end{lemma}
We now present the error bound for the matrix Dantzig selector as was obtained in ~\cite{Candes2010} in Theorem ~\ref{theorem:Dantzig_recovery}. In Corollary ~\ref{corollary:rank_k_approximate}, we exploit this result in our setting for $r=k$ in order to obtain the error bound for recovering the rank-$k$ approximation $\widehat{\matX}^{(k)}_{DS}$ to $\matX$.
%
\begin{theorem} \label{theorem:Dantzig_recovery}
Let \text{rank}($\matX$) $\leq$ $r$ and let $\widehat{\matX}_{DS}$ be the solution to \eqref{eq:MDS}. If $\kappa_{4r} <\kappa < \sqrt{2}-1$ and $\norm{\Phi^{*}(\vecnoise)} \leq \lambda$, then we have with probability at least $1-2e^{-m_{\Phi}q(\kappa) + 4r(d+m_{\setX}+1)u(\kappa)}$ that
\begin{equation*}
\norm{\widehat{\matX}_{DS}-\matX}_F^2 \leq C_0r\lambda^2,
\end{equation*}
where $C_0$ depends only on the isometry constant $\kappa_{4r}$.
\end{theorem}
%
\begin{corollary} \label{corollary:rank_k_approximate}
Denoting $\widehat{\matX}_{DS}$ to be the solution of ~\eqref{eq:MDS}, if $\widehat{\matX}^{(k)}_{DS}$ is the best rank-$k$ approximation to $\widehat{\matX}_{DS}$ in the sense of $\norm{\cdot}_F$, and if $\kappa_{4k} < \kappa < \sqrt{2}-1$, then we have
\begin{equation*}
\norm{\matX - \widehat{\matX}^{(k)}_{DS}}_F^2 \leq \frac{C_0 C_2^2 k^5 \epsilon^2 d^2 m_{\setX}^2}{m_{\Phi}}(1+\kappa),
\end{equation*}
with probability at least $1-2e^{-m_{\Phi}q(\kappa) + 4k(d+m_{\setX}+1)u(\kappa)}$, where the constant $C_0$ depends only on $\kappa_{4k}$.
\end{corollary}

Corollary \ref{corollary:rank_k_approximate} is the main result of this subsection, which is proved in Appendix \ref{sec: ap: cor4}. 
%
\subsection{Approximation of $\matA$}{\label{subsec:approx_A}} In the previous subsection, we derive a rank-$k$ approximation $\widehat{\matX}^{(k)}_{DS}$ of the original rank-$k$ matrix $\matX$ with a bound on the approximation error $\norm{\widehat{\matX}^{(k)}_{DS} - \matX}_F$. Here, we are interested in recovering an approximation $\widehat{\matA}$ to the matrix $\matA$ from $\widehat{\matX}^{(k)}_{DS}$. Trivially, this can be achieved by setting $\widehat{\matA}$ to the left singular vector matrix of $\widehat{\matX}^{(k)}_{DS}$. The purpose of the analysis here is to theoretically characterize the ensuing approximation error.

Let the SVD of $\matX$ and $\widehat{\matX}^{(k)}_{DS}$ be
$\matX = \matA^T \matG = \matA^T \matU_G \Sigma_G \matV_G^T = \matA_1^T\Sigma_G \matV_G^T$ and $\widehat{\matX}^{(k)}_{DS} = \widehat{\matA}^T \widehat{\Sigma} \widehat{\matV}$, respectively. Then, $\Sigma = \text{diag}(\sigma_1,\sigma_2,\dots,\sigma_k) \quad \text{and} \quad \widehat{\Sigma} = \text{diag}(\widehat{\sigma}_1, \widehat{\sigma}_2,\dots, \widehat{\sigma}_k)$ are diagonal matrices with $\sigma_1 \geq \sigma_2 \geq \dots \sigma_k$ and $\widehat{\sigma}_1 \geq \widehat{\sigma}_2 \geq \dots \widehat{\sigma}_k$, respectively. Moreover, $\matU_G$ is a $k \times k$ unitary matrix. The columns of $\matA_1^T,\matV_G^T$ and $\widehat{\matA},\widehat{\matV}$ are the singular vectors of $\matX$ and $\widehat{\matX}^{(k)}_{DS}$, respectively. Finally, we have $\sigma_i = \sqrt{\lambda_i(\matG\matG^T)}$ where $\lambda_i$ denotes the $i^{th}$ eigenvalue of
\begin{equation} \label{eq:matrix_sum}
\matG\matG^T = \sum_{j=1}^{m_{\setX}}\left(\nabla g(\matA \vecxij) \nabla g(\matA \vecxij)^T \right).
\end{equation}
We now show that if $\norm{\matX-\widehat{\matX}^{(k)}_{DS}}_F$ is driven to be smaller than a threshold then it leads to a probabilistic lower bound on $\norm{\matA\widehat{\matA}^T}_F$. Lemma~\ref{lemma:mata_approximation}, proved in Appendix \ref{sec:proof_lemma_mata_approx}, precisely states this fact.
%
\begin{lemma} \label{lemma:mata_approximation}
For a fixed $0 < \rho < 1$, $m_{\setX} \geq 1$, $m_{\Phi} < m_{\setX}d$ if
$
\epsilon < \frac{\displaystyle 1}{\displaystyle C_2 k^2 d (\sqrt{k}+\sqrt{2})}\left(\frac{\displaystyle (1-\rho) m_{\Phi}\alpha}{\displaystyle (1+\kappa) C_0 m_{\setX}}\right)^{1/2},
$
then with probability at least
$
1 - k\exp\left\{-\frac{m_{\setX}\alpha \rho^2}{2 k C_2^2}\right\} - 2\exp\left\{-m_{\Phi}q(\kappa) + 4k(d+m_{\setX}+1)u(\kappa)\right\}
$
we have
\begin{equation*}
\norm{\matA\widehat{\matA}^T}_F \geq \left(k - \frac{2\tau^2}{(\sqrt{(1-\rho)m_{\setX}\alpha} - \tau)^2} \right)^{1/2},
\end{equation*}
where $\tau^2 = \frac{\displaystyle C_0 C_2^2 k^5 \epsilon^2 d^2 m_{\setX}^2}{\displaystyle m_{\Phi}}(1+\kappa)$ is the error bound derived in Corollary ~\ref{corollary:rank_k_approximate}.
\end{lemma}

\paragraph{\bf Choice of $\epsilon$.} We note here that a guaranteed lower bound on $\norm{\matA\widehat{\matA}^T}_F$, of say $(k\eta)^{1/2}$ for some $0 < \eta < 1$, follows along the lines of the proof in Appendix \ref{sec:proof_lemma_mata_approx} by ensuring that the following holds:
\begin{equation*}
\epsilon < \frac{\displaystyle 1}{\displaystyle C_2 k^2 d (\sqrt{k(1-\eta)}+\sqrt{2})}\left(\frac{\displaystyle (1-\rho) m_{\Phi}\alpha(1-\eta)}{\displaystyle (1+\kappa) C_0 m_{\setX}}\right)^{1/2}.
\end{equation*}

%
\subsection{Approximation of $f$}{\label{subsec:approx_f_gen}}
We now have the necessary background to state our main approximation result for
the function $f$.
%
\begin{theorem}{(Main approximation theorem)} \label{thm:main_approx_thm}
Let us fix $\delta \in \Real^{+}$, $0 < \rho < 1, 0 < \kappa < \sqrt{2}-1$.
Under the assumptions and notations mentioned earlier, for a fixed $m_{\setX}
\geq 1$, $m_{\Phi} < m_{\setX}d$ and $\epsilon < \frac{\displaystyle
\delta}{\displaystyle C_2 k^{5/2} d (\delta +
2C_2\sqrt{2k})}\left(\frac{\displaystyle (1-\rho) m_{\Phi}\alpha}{\displaystyle
(1+\kappa) C_0 m_{\setX}}\right)^{1/2}$ we have that the function
$\widehat{f}(\vecx) = \widehat{g}(\widehat{\matA}\vecx)$ defined by means of
$\widehat{g}(y):= f(\widehat{\matA}^Ty), \quad \vecy \in B_{\Real^k}(1+\bareps)$
has the uniform approximation bound
\begin{equation*}
\norm{f-\widehat{f}}_{\infty} \leq \delta,
\end{equation*}
with probability at least $ 1 - k\exp\left\{-\frac{m_{\setX}\alpha \rho^2}{2 k
C_2^2}\right\} - 2\exp\left\{-m_{\Phi}q(\kappa) +
4k(d+m_{\setX}+1)u(\kappa)\right\}.
$
\end{theorem}
%
We provide the proof of our main approximation result Theorem
\ref{thm:main_approx_thm} in Appendix \ref{sec: ap: proof of main theorem}. In Section \ref{sec:tract_samp_complexity}, we establish the tractability
of our learning algorithm and also provide a comparison of our sampling bounds with those of \cite{Fornasier2010} (i.e. $\matA$ is compressible) for
different function classes. In particular, we show that our sampling bounds can be better than \cite{Fornasier2010} depending on the compressibility of $\matA$. 
For instance, if $1<q<2$, then our bounds exhibit better scaling. Furthermore the results of \cite{Fornasier2010} also benefit from our proposition that shows how the
parameter $\alpha$ behaves for a variety of models such as the class of additive function models.
\begin{remark}
(i) We can also consider approximating functions of the form: $f(\vecx) =
g(\matA\vecx + \vecb)$, assuming without loss of generality that
$\norm{\vecb}_{\ell_2^k} \leq 1$. Then, our estimator $\widehat{f}$ attains the
following form: $\widehat{f}(\vecx) = \widehat{g}(\widehat{\matA}\vecx) =
g(\matA\widehat{\matA}^T\widehat{\matA}\vecx + \vecb)$, where $\widehat{g}(y):=
f(\widehat{\matA}^Ty), \ \vecy \in B_{\Real^k}(1+\bareps)$. It is
straightforward to verify that we obtain the same approximation bound on
$\norm{f-\widehat{f}}_{\infty}$ along the lines of the proof of Theorem
~\ref{thm:main_approx_thm}. Furthermore, we can then uniformly approximate the
function $\hat{g}$ by first sampling it on a rectangular grid  $h\mathbb{Z}^k
\cap (-(2+\bar{\epsilon}),(2+\bar{\epsilon}))^k$ as before with uniformly spaced
points in each direction. Subsequently, by using quasi interpolants to
interpolate between the points we obtain an approximation $\hat{g}_h$. In this
particular setting, we need not approximate $\vecb$ to derive approximation
guarantees on $f$. In particular we need only use a bound on
$\norm{\vecb}_{\ell_2^k}$ to accordingly set the size of the sampling grid. \newline

\noindent (ii) In Theorem \ref{thm:main_approx_thm} the step size parameter $\epsilon$ needs to be suitably small in order 
to guarantee the approximation result on $f$. This suugests that for large $d$, the requirement on $\epsilon$ might be too strict leading to numerical issues
in approximating the gradient of $f$ by finite differences as in \eqref{eq: first order ridge relation}. However note  
that the bound on $\epsilon$ depends on the ratio $\sqrt{m_{\Phi}/m_{\setX}}$. Hence one can also choose a constant $\epsilon$
and $m_{\setX} = O(1/\alpha)$. We can then compensate the choice of $\epsilon$
by choosing a suitably large value of $m_{\Phi}$ (as determined from Theorem \ref{thm:main_approx_thm} by the parameters $k,d,C_2,C_0,\rho,\kappa,\delta$ and $\alpha$)
resulting in a good approximation to $f$ with high probability. We also note in our numerical simulations in Section \ref{sec: presim} that it suffices to consider 
reasonable values such as $\epsilon \sim 10^{-3}$ which leads to stable approximation results.
\end{remark} 
%
\subsection{Impact of measurement noise on learning scheme} \label{subsec:imp_meas_noise}
For the simplicity of our subsequent theoretical analysis, we fix $\epsilon$ as a small constant. As a by-product, $\epsilon^{-1}$ linearly amplifies the oracle Gaussian noise within the perturbation model \eqref{eq: noisy oracle}. This is inherently due to the way we leverage the oracle calls while forming our naive gradient estimates: ${\epsilon}^{-1}\left(f(\vecx+\epsilon\phi)- f(\vecx)\right)$. We note, however, that there are much better ways in practice to exploit the noisy oracle values to obtain de-noised gradient estimates by adaptively varying the region size and collectively using the oracle values (e.g., in the manner of regression methods in statistics or trust-region methods in optimization). Of course, the ideal solution in our formulation is to have access to a \emph{gradient oracle}, which has small perturbations. We now further address these issues here.

\paragraph{\bf Gaussian noise.} Let us first assume that the evaluation of $f$ at a point $\vecx \in B_{\Real^d}(1+\bareps)$ yields: $f(\vecx) + Z$, where $Z \sim \mathcal{N}(0,\sigma^2)$. Thus under this noise model, ~\eqref{factorization_equation_gen} changes to:
\begin{equation} \label{eq:noisy_eq}
\Phi(\matX) = \vecy + \vecnoise + \vecz
\end{equation}
where $\vecz \in \Real^{m_{\Phi}}$ and $z_i = \sum_{j=1}^{m_{\setX}} \frac{\displaystyle z_{ij}}{\displaystyle \epsilon}$. Assuming the iid noise samples, we have $z_{ij} \sim \mathcal{N}(0,2\sigma^2)$, and $
z_i \sim \mathcal{N}\left(0,\frac{2 m_{\setX} \sigma^2}{\epsilon^2}\right)$ for $ i = 1,\dots,m_{\Phi}.$ Therefore, the noise variance gets amplified by a polynomial factor $\frac{m_{\setX}}{\epsilon^2}$.

In our analysis, the parameter $\epsilon$ is assumed to be sufficiently small. In fact, Lemma ~\ref{lemma:mata_approximation} requires
\begin{equation*}
\epsilon < \frac{\displaystyle 1}{\displaystyle C_2 k^2 d (\sqrt{k}+\sqrt{2})}\left(\frac{\displaystyle (1-\rho) m_{\Phi}\alpha}{\displaystyle (1+\kappa) C_0 m_{\setX}}\right)^{1/2}.
\end{equation*}
Therefore, for large $d$, $\epsilon$ is  at most $\mathcal{O}\left(\frac{\displaystyle \alpha^{1/2}}{\displaystyle d}\right)$. To make the matters worse, the next section shows that $\alpha$ can be at most $\mathcal{O}(1)$ and usually decays polynomially with $d$. Thus, we see that the noise variance gets amplified as the dimension $d$ and the number of samples $m_{\setX}$ increases.

To further elaborate on how this affects the low rank recovery scheme, recall that in the convex program ~\eqref{eq:MDS}, we require the true matrix $\matX$ to be feasible. In the setting of ~\eqref{eq:noisy_eq}, this behooves us to consider
$\norm{\Phi^{*}(\vecnoise + \vecz)} \leq \lambda
$ for the feasibility of the solution. Let $m = \text{max}(m_{\Phi} , m_{\setX})$. Then, Lemma 1.1 ~\cite{Candes2010} leads to the following bound with high probability ($\gamma > 2\sqrt{\log 12}$)
\begin{equation*}
\norm{\Phi^{*}(\vecz)} \leq 2\gamma\sqrt{(1 + \kappa_1)m}\sqrt{\frac{2 m_{\setX} \sigma^2}{\epsilon^2}}
\end{equation*}
Using this with result of Lemma ~\ref{lemma:adjoint_noise_bound}, the following bound holds with high probability for $\gamma > 2\sqrt{\log 12}$
\begin{equation*}
\norm{\Phi^{*}(\vecnoise + \vecz)} \leq \frac{2\gamma\sigma}{\epsilon}\sqrt{2m(1+\kappa_1)m_{\setX}} + \frac{\displaystyle C_2 \epsilon d m_{\setX} k^2}{\displaystyle 2\sqrt{m_{\Phi}}}(1+\kappa_1)^{1/2}.
\end{equation*}

We observe that as opposed to the perfect oracle setting we can no longer control the upper bound on $\norm{\Phi^{*}(\vecnoise+\vecz)}$ by simply reducing $\epsilon$, due to the appearance of the ($1 / \epsilon$) term. Hence, unless $\sigma$ is $\mathcal{O}(\epsilon)$ or less, (e.g., $\sigma$ reduces with $d$), we can declare that our learning scheme with the matrix Dantzig selector is sensitive to noise, also when we use the minimum number of samples for recovery and we do not change the way we calculate the gradients. However, in many practical cases, it is possible to increase the number samples by a factor of $d$ since noisy oracles tend to be cheaper. Alternatively, we must leverage the noisy oracle samples with more sophisticated methods to obtain denoised gradient estimates. Hence, for additional stability against Gaussian oracles with a constant noise variance, our tractability results in Section \ref{sec:tract_samp_complexity} needs to multiplied by a polynomial factor of $d$.

%
\section{Information Complexity of Oracle-based Low-Rank Learning} \label{sec:tract_samp_complexity}
In this section, we establish the tractability of our approximation strategy. As the first step, we note that the uniform approximation result in Theorem~\ref{thm:main_approx_thm} holds with probability $1-p_1-p_2$ when
\begin{equation}
m_{\setX} > \frac{\displaystyle 2kC_2^2}{\displaystyle \alpha\rho^2}\log(k/p_1), \quad m_{\Phi} > \frac{\displaystyle \log(2/p_2)+4k(d+m_{\setX}+1)u(\kappa)}{\displaystyle q(\kappa)}. \label{eq:samp_complex}
\end{equation}
Therefore, for a desired probability of success, the sampling complexities scales as $m_{\setX} = \mathcal{O}\left(\frac{\displaystyle k\log k}{\displaystyle \alpha}\right)$ and $m_{\Phi} = \mathcal{O}(k(d+m_{\setX}))$ 
for large  $d$. At this juncture, while we seemingly have the complexity of our randomized sampling scheme in Section \ref{sec: sampling}, the effect of the parameter $\alpha$ is still implicit.

Appendix \ref{subsec:tractability_gen} relates the parameter $\alpha$ to the Hessian matrix $H^f$ in our Ansatz in Section \ref{sec: problem_setup}. 
Based on this discussion, we can rigorously observe that the conditioning of the matrix $H^f$ for large $d$ would be determined predominantly by the behavior of $g$ in a open neighborhood around the origin. 
This behavior is quite straightforward to analyze when $k=1$. What is not so easy to characterize is the behavior when $k>1$. For instance, \cite{Fornasier2010} finds 
it necessary to further constrain $f$ to be a radial function to analyze the behavior of $\alpha$ when $k>1$. By radial function, we mean 
$f(\vecx) = g(\matA\vecx) = g_0(\norm{\matA\vecx}_{l_2^{k}})$, where $g_0$ is $\mathcal{C}^2$ smooth due to our problem set up.

One of the main contributions in this work is that we provide a local condition in Proposition ~\ref{prop:alpha_tract_gen} below (proved in Appendix \ref{sec: ap: pprop2}) 
that alleviates required conditions on the global structure of $f$:

\begin{prop} \label{prop:alpha_tract_gen}
Assume that $g\in\mathcal{C}^2:B_{\Real^k} \rightarrow \Real$ has Lipschitz continuous second order partial derivatives in an open neighborhood of the origin, 
$\mathcal{U}_{\theta} = B_{\Real^k}(\theta)$ for some fixed $\theta$ (depending only on $k$ with $k$ fixed):
\begin{equation*}
\frac{\abs{\frac{\displaystyle \partial^2 g}{\displaystyle \partial y_i \partial y_j}(\vecy_1) - 
\frac{\displaystyle \partial^2 g}{\displaystyle \partial y_i \partial y_j}(\vecy_2)}}{\norm{\vecy_1 - \vecy_2}_{l_2^k}} < L_{i,j} \quad 
\forall \vecy_1,\vecy_2 \in \mathcal{U}_{\theta}, \vecy_1 \neq \vecy_2, \ i,j= 1,\dots,k.
\end{equation*}
Denoting $L = \max_{1 \leq i,j \leq k} L_{i,j}$, assume that $\nabla^2 g(\veco)$ is full rank, and either one of the following conditions hold:  
\begin{enumerate}
\item $\nabla g(\veco) = \veco$.

\item $\nabla g(\veco) \neq \veco$ and $L = O(1/d)$.
\end{enumerate}
Then, we have $\alpha = \Theta(1/d)$ as $d \rightarrow \infty$.
\end{prop}

We are now ready to consider example function classes for $k=1$ as well as $k>1$ below, and derive the sampling complexities. 
As a baseline, we compare each result with \cite{Fornasier2010} to highlight the variations as a result of forgoing the compressibility assumption on $A$.
%

%
\subsection{Function classes for $k = 1$}
\cite{Fornasier2010} defines the following sets of classes of $\mathcal{C}^2$ smooth ridge functions for the case $k = 1$, 
for which they establish the scaling behavior of $\alpha$ to be polynomial in $1/d$:
\begin{enumerate}

\item~[$0 < q < 1$, $C_1 > 1$ and $C_2 \geq \alpha_0 > 0$]: $\mathcal{F}_d^{1} := \mathcal{F}_d^{1} (\alpha_0,q,C_1,C_2):= \{f: B_{\Real^d} \rightarrow \Real |
\exists \veca \in \Real^d, \norm{\veca}_{\ell_2^d} = 1, \norm{\veca}_{\ell_q^d} \leq C_1 $ {and} $
\exists g \in \mathcal{C}^2(B_{\Real}) , \abs{g^{\prime}(\veco)} \geq \alpha_0 > 0 : f(\vecx) = g(\veca^T\vecx)\}. $


\item~[For an open neighborhood $\mathcal{U}$ of 0, $0 < q < 1$, $C_1 > 1$, $C_2 \geq \alpha_0 > 0$ and $M \in \mathbb{N}$]: $
\mathcal{F}_d^{2} := \mathcal{F}_d^{2} (\mathcal{U},\alpha_0,q,C_1,C_2,M):= \{f: B_{\Real^d} \rightarrow \Real :
\exists \veca \in \Real^d, \norm{\veca}_{l_2^d} = 1, \norm{\veca}_{l_q^d} \leq C_1$ and
$ \exists g \in \mathcal{C}^2(B_{\Real}) \bigcap \mathcal{C}^{M+2}(\mathcal{U}), g^{(N)}(\veco) = 0 \quad \forall \quad 1 \leq N  \leq M ,$ $
\abs{g^{(M+1)}(\veco)} \geq \alpha_0 > 0 : f(\vecx) = g(\veca^T\vecx)\}.
$
\end{enumerate}

We now generalize the above two classes in two non-trival ways:
\begin{enumerate}
\item {By doing away with the compressibility assumption on $\veca$ from both $\mathcal{F}_d^{1}$ and $\mathcal{F}_d^{2}$.}

\item {By showing along the lines of the proof of Proposition ~\ref{prop:alpha_tract_gen} that in $\mathcal{F}_d^{2}$, one can relax the space: 
$\mathcal{C}^2(B_{\Real}) \bigcap \mathcal{C}^{M+2}(\mathcal{U})$ to 
$\mathcal{C}^2(B_{\Real}) \bigcap \mathcal{C}^{M+1}(\mathcal{U}) \bigcap \mathcal{L}^{M+1}(\mathcal{U}, L)$. Here $\mathcal{L}^{M+1}(\mathcal{U}, L)$ 
denotes the space of $\mathcal{C}^{M+1}(\mathcal{U})$ functions whose $(M+1)^{th}$ derivatives are Lipschitz continuous with constant $L$.}
\end{enumerate}

For the sake of completeness, here are our generalized function classes:
\begin{enumerate}
\item~[$C_2 \geq \alpha_0 > 0$]: $\mathcal{H}_d^{1} := \mathcal{H}_d^{1} (\alpha_0,C_2):= \{f: B_{\Real^d} \rightarrow \Real |
\exists \veca \in \Real^d, \norm{\veca}_{l_2^d} = 1,$ and $\exists g \in \mathcal{C}^2(B_{\Real}), \abs{g^{\prime}(\veco)} \geq \alpha_0 > 0 : f(\vecx) = g(\veca^T\vecx)\}.$


\item~[For an open neighborhood $\mathcal{U}$ of 0, $C_2 \geq \alpha_0 > 0$, $0 < L < \infty$ and $M \in \mathbb{N}$]:
$ \mathcal{H}_d^{2} := \mathcal{H}_d^{2} (\mathcal{U},\alpha_0,C_2,M,L):= \{f: B_{\Real^d} \rightarrow \Real :
\exists \veca \in \Real^d, \norm{\veca}_{l_2^d} = 1 $ and $\exists g \in \mathcal{C}^2(B_{\Real}) \bigcap \mathcal{C}^{M+1}(\mathcal{U}) $ $ \bigcap \mathcal{L}^{M+1}(\mathcal{U}, L),
g^{(N)}(\veco) = 0 \quad \text{for all} \quad 1 \leq N  \leq M,$ $
\abs{g^{(M+1)}(\veco)} \geq \alpha_0 > 0 : f(\vecx) = g(\veca^T\vecx)\}.
$
\end{enumerate}

Table ~\ref{tab:samp_compl_k_1} summarizes the sampling complexities for the above function classes. Observe that the sampling complexity increases 
from $\mathcal{O}(\log d)$ to $\mathcal{O}(d)$ when $g^{\prime}(0) \neq 0$ and from $\mathcal{O}(d^{\frac{2M}{2-q}})$ to $\mathcal{O}(d^{2M})$ when the first $M$ order partial derivatives of $g$ at the origin are 0.

%
\begin{table}[!htp]
\centering 
\begin{tabular}{| c | c | c | c | c |} 
\hline 
Function class & Scaling of $\alpha$ & $m_{\setX}$ & $m_{\Phi}$ & $m_{\setX} \times (m_{\Phi} + 1)$ \\[1.5pt]
\hline $\mathcal{F}_d^{1}$ & $\Theta(1)$ & $\mathcal{O}(1)$ & $\mathcal{O}(\log d)$ & $\mathcal{O}(\log d)$ \\[1.5pt]
       $\mathcal{H}_d^{1}$ & $\Theta(1)$ & $\mathcal{O}(1)$ & $\mathcal{O}(d)$ & $\mathcal{O}(d)$ \\ \hline\hline
       $\mathcal{F}_d^{2}$ & $\Theta(d^{-M})$ & $\mathcal{O}(d^M)$ & $\mathcal{O}\left(d^{\frac{Mq}{2-q}}\right)$ & $\mathcal{O}\left(d^{\frac{2M}{2-q}}\right)$ \\[1.5pt]
       $\mathcal{H}_d^{2}$ & $\Theta(d^{-M})$ & $\mathcal{O}(d^M)$ & $\mathcal{O}(d^M)$ & $\mathcal{O}(d^{2M})$ \\
\hline
\end{tabular}
\newline
\caption[Short]{\small Comparison of sampling complexities for approximating $f$ when $\veca$ is compressible (function classes $\mathcal{F}_d^{1}$, $\mathcal{F}_d^{2}$) 
with those when no compressibility assumption is made on $\veca$ (function classes $\mathcal{H}_d^{1}$, $\mathcal{H}_d^{2}$).}\label{tab:samp_compl_k_1}
\end{table}

%
\subsection{Function classes for $k > 1$}
The case $k>1$ is significantly more challenging to handle as compared to the case $k = 1$. \cite{Fornasier2010} shows that if $f$ is a radial function, 
$f(\vecx) = g(\matA\vecx) = g_0(\norm{\matA\vecx}_{l_2^{k}})$, where $g_0$ is $\mathcal{C}^2$, then they can handle the following scenario depending on the local smoothness properties of $g_0$:

[For an open neighborhood $\mathcal{U}$ of 0]: $
\mathcal{G}_{d,k} := \{
M \in \mathbb{N}, g_0 \in \mathcal{C}^2(B_{\Real}) \bigcap \mathcal{C}^{M+2}(\mathcal{U}), g_0^{(N)}(0) = 0 \quad \forall  1 \leq N \leq M$ and $\abs{g_0^{(M+1)}(0)} \geq \alpha_0 > 0\}$.

In particular the authors show that for the above function class, $\alpha = \Theta(d^{-M})$. The proof of this result can be found in Section 
4.3 of ~\cite{Fornasier2010}. Table ~\ref{tab:sampl_compl_k_rad} provides a comparison of sampling complexities between ~\cite{Fornasier2010} and our work for the function class $\mathcal{G}_{d,k}$.
%
\begin{table}[!htp]
\centering
\begin{tabular}[!htp]{| c | c | c | c | c |} 
%
\hline
$g_0 \in \mathcal{G}_{d,k}$ & Scaling of $\alpha$ &$m_{\setX}$ & $m_{\Phi}$ &  $m_{\setX} \times (m_{\Phi} + 1)$ \\[2.5pt]
\hline

Compressible $\matA$ & $\Theta(d^{-M})$ & $\mathcal{O}(k d^M \log k)$ & $\mathcal{O}\left(k^{\frac{2}{2-q}} d^{\frac{Mq}{2-q}}\right)$ & $\mathcal{O}\left(k^{\frac{4-q}{2-q}}d^{\frac{2M}{2-q}}\log k\right)$ \\[3pt]

Arbitrary $\matA$ & $\Theta(d^{-M})$ &$\mathcal{O}(k d^M \log k)$ & $\mathcal{O}(k^2 d^M \log k)$ & $\mathcal{O}(k^3 d^{2M} (\log k)^2)$ \\[2.5pt]
\hline
\end{tabular}
\caption[Short]{\small Comparison of sampling complexities for approximating radial functions: $f(\vecx) = g_0(\norm{\matA\vecx}_{l_2^{k}})$.}
\label{tab:sampl_compl_k_rad}
\end{table}
\begin{remark}
Note that in function class denoted by $\mathcal{G}_{d,k}$, we require $g^{\prime}_0(0) = 0$, since otherwise $g(\cdot)$ would not be differentiable at the origin.
\end{remark}

We now qualitatively demonstrate our generalization of the above function class via our Proposition ~\ref{prop:alpha_tract_gen} and highlight its significance. 
Assume that $f(\vecx) = g(\matA\vecx)$ where $g$ has the following form:
\begin{equation}
g(y_1,\dots,y_k) = \sum_{l=1}^{k} g_l(y_l). \label{eq:eg_additive_g}
\end{equation}

\noindent We have $\frac{\displaystyle \partial g}{\displaystyle \partial y_i} = g^{\prime}_i (y_i)$ and, 
$\nabla^2 g(\vecy) = \text{diag}(g_1^{\prime\prime}(y_1),\dots,g_k^{\prime\prime}(y_k))$. Clearly, $\nabla^2 g(\veco)$ is full rank if and only if $g_i^{\prime\prime}(0) \neq 0 \ \forall \ i=1,\dots,k$. 
Hence, we conclude that if the individual $g_i$'s in ~\eqref{eq:eg_additive_g} are such that for each $i=1,\dots,k$, we have $g_i^{\prime\prime}(0) \neq 0$, and $g_i^{\prime\prime}$ is 
Lipschitz continuous in an open neighborhood of the origin, then the function $g$ would satisfy the conditions of Proposition ~\ref{prop:alpha_tract_gen} resulting in $\alpha = \Theta(1/d)$ for large $d$. 
To give a few practical examples of such $g_i$'s one could think of smooth kernel functions such as Gaussian and Epanechnikov, kernels used commonly in non-parametric estimation ~\cite{Li2007}. 
Furthermore, the sample complexity for learning functions belonging to the class specified by Proposition \ref{prop:alpha_tract_gen} can be seen from Table \ref{tab:sampl_compl_k_rad} by setting 
$M = 1$ (since $\alpha = \Theta(1/d)$). Thus the sample complexity for abitrary $\matA$ is $\mathcal{O}(k^3 d^{2} (\log k)^2)$, while for compressible $\matA$ it is 
$\mathcal{O}\left(k^{\frac{4-q}{2-q}}d^{\frac{2}{2-q}}\log k\right)$.

\begin{remark}
One can think of extending the conditions of Proposition ~\ref{prop:alpha_tract_gen} so that the first $M$ order partial derivatives are 0. However, we choose to 
restrict our analysis to $\mathcal{C}^2$ smooth ridge functions obeying the variation conditions as defined in Proposition ~\ref{prop:alpha_tract_gen} as it 
enables us to state conditions on the Hessian of $g$ evaluated at the origin which is more intuitive to interpret and easy to verify.
\end{remark}

\section{Numerical Experiments}\label{sec: presim}
We present simulation results for functions of the form $f(\vecx) = g(\matA\vecx)$ with $\matA$ being the linear parameter matrix. We assume $\matA$ to be row orthonormal and concern ourselves only with the recovery of $\matA$ upto an orthonormal transformation.
%
\subsection{Logistic function ($k = 1$)}
We first take $k=1$ and consider $f(\vecx) = g(\veca^T \vecx)$ where $g$ is the logistic function:
\begin{equation*}
g(y) = \frac{1}{1 + e^{-y}}.
\end{equation*}
One can easily verify that $C_2 = \sup_{\abs{\beta} \leq 2} \abs{g^{(\beta)}(y)} = 1$. Furthermore we compute the value of $\alpha$ through the following approximation, which holds for large d:
\begin{equation*}
\alpha = \int\abs{g^{\prime}(\veca^T \vecx)}^2 d\mu_{\mathbb{S}^{d-1}} \approx \abs{g^{\prime}(0)}^2 = (1/16).
\end{equation*}

We require $\abs{\langle \hat{\veca},\veca \rangle}$ to be greater then 0.99. We fix values of $\kappa < \sqrt{2}-1$, $\rho \in (0,1)$ and $\epsilon = 10^{-3}$. The value of $m_{\setX}$ (number of points sampled on $\mathbb{S}^{d-1}$) is fixed at 20 and we vary $d$ over the range 200-3000. For each value of $d$, we increase $m_{\Phi}$ till $\abs{\langle \hat{\veca},\veca \rangle}$ reaches the specified performance criteria. We remark that for each value of $d$ and $m_{\Phi}$, we choose $\epsilon$ to satisfy the bound in Lemma ~\ref{lemma:mata_approximation} for the specified performance criteria given by $\eta$.

\begin{figure}[htp]
\centering
\includegraphics[scale = 0.5]{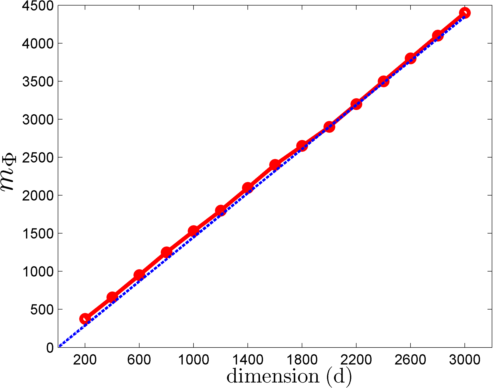}
\caption{\small{Plot of $\frac{m_{\Phi}}{d}$ versus $d$ for $m_{\setX} = 20$ , with $m_{\Phi}$ chosen to be minimum value needed to achieve $\abs{\langle \hat{\veca},\veca \rangle} \ \geq \ 0.99$. $\epsilon$ is fixed at $10^{-3}$. $m_{\Phi}$ scales approximately linearly with $d$ with a scaling factor around 1.45.}} \label{fig:ridge_logistic}
\end{figure}

Figure \ref{fig:ridge_logistic} depicts the scaling of $m_{\Phi}$ with the dimension $d$. The results are obtained by selecting $\veca$ uniformly at random on $\mathbb{S}^{d-1}$ and averaging the value of $\abs{\langle \hat{\veca},\veca \rangle}$ over 10 independent trials. We observe that for large values of $d$, the minimum number of directional derivatives needed to achieve the performance bound on $\abs{\langle \hat{\veca},\veca \rangle}$ scales approximately linearly with $d$, with a scaling factor of around 1.45.

%
\subsection{Sum of Gaussian functions ($k > 1$)}
We next consider functions of the form $f(\vecx) = g(\matA\vecx + \vecb) = \sum_{i=1}^{k} g_i(a_i^T\vecx + b_i)$, where:
\begin{equation*}
g_i(y) = \frac{1}{\sqrt{2\pi\sigma_i^2}}\exp\left(-\frac{(y+b_i)^2}{2\sigma_i^2}\right)
\end{equation*}

\begin{figure}[htp]
\centering
\includegraphics[scale = 0.5]{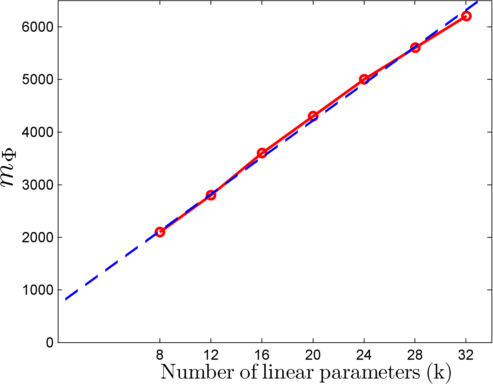}
\caption{\small{Plot of $m_{\Phi}$ versus $k$ for $d = 100, m_{\setX} = 100$ , with $m_{\Phi}$ chosen to be minimum value needed to achieve $\frac{1}{k}\norm{\matA\widehat{\matA}}^2_F \ \geq \ 0.99$.}} \label{fig:ridge_gaussian}
\end{figure}

We fix $d=100$, $\epsilon= 10^{-3}$, $m_{\setX} = 100$ and vary $k$ from 8 to 32 in steps of 4. For each value of $k$ we are interested in the minimum value of $m_{\Phi}$ needed to achieve $\frac{1}{k}\norm{\matA\widehat{\matA}}^2_F \ \geq \ 0.99$. In Figure ~\ref{fig:ridge_gaussian} we see that $m_{\Phi}$ scales approximately linearly with the number of gaussian atoms, $k$. The results are averaged over 10 trials. In each trial, we select the rows of $\matA$ over the left Haar measure on $\mathbb{S}^{d-1}$, and the parameter $\vecb$ uniformly at random on $\mathbb{S}^{k-1}$ scaled by a factor 0.2. Furthermore we generate the standard deviations of the individual Gaussian functions uniformly over the range [0.1 0.5].

%
\subsection{Impact of Noise}
\begin{figure}[htp]
\centering
\includegraphics[scale = 0.5]{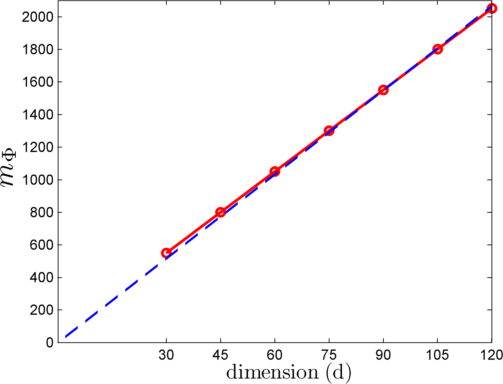}
\caption{\small{Plot of $m_{\Phi}$ versus $d$ for $k = 5, m_{\setX} = 30$ , with $m_{\Phi}$ chosen to be minimum value needed to achieve $\frac{1}{k}\norm{\matA\widehat{\matA}}^2_F \ \geq \ 0.99$. Each point query is corrupted with Gaussian noise of standard deviation: $0.01/d^{3/2}$.}} \label{fig:ridge_quad_gauss_noise}
\end{figure}

We now consider quadratic forms, i.e. $f(\vecx) = g(\matA\vecx) = \norm{\matA\vecx - b}^2$ with the point queries corrupted with Gaussian noise. Since for $g(\vecy) = \norm{y-\vecb}^2$ we have $\nabla^2 g(\vecb)$ to be full rank diagonal, we take $\alpha$ to be $1/d$. We fix $k = 5$, $m_{\setX} = 30$, $\epsilon = 10^{-1}$ and vary $d$ from 30 to 120 in steps of 15. For each $d$ we perturb the point queries with Gaussian noise of standard deviation: $0.01/d^{3/2}$. This is the same as repeatedly sampling each random location approximately $d^{3/2}$ times followed by averaging. We then compute the minimum value of $m_{\Phi}$ needed to achieve $\frac{1}{k}\norm{\matA\widehat{\matA}}^2_F \ \geq \ 0.99$. We average the results over 10 trials, and in each trial, we select the rows of $\matA$ over the left Haar measure on $\mathbb{S}^{d-1}$. The parameter $\vecb$ is chosen uniformly at random on $\mathbb{S}^{k-1}$. In Figure \ref{fig:ridge_quad_gauss_noise} we see that $m_{\Phi}$ scales approximately linearly with $d$.

\begin{figure}[htp]
\centering
\includegraphics[scale = 0.5]{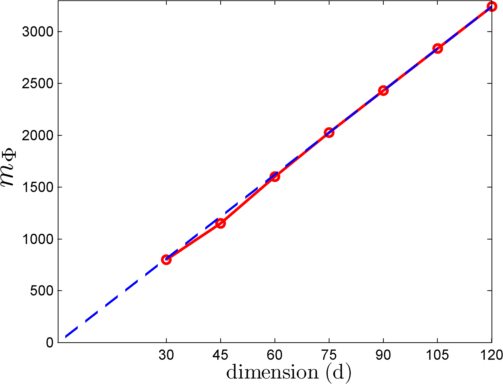}
\caption{\small{Plot of $m_{\Phi}$ versus $d$ for $k = 5, m_{\setX} = 30$ , with $m_{\Phi}$ chosen to be minimum value needed to achieve $\frac{1}{k}\norm{\matA\widehat{\matA}}^2_F \ \geq \ 0.99$. With probability $0.01$, each point query is corrupted with Gaussian noise of standard deviation: $0.01$.}} \label{fig:ridge_quad_sparse_noise}
\end{figure}

We next repeat the above experiment under a different noise model. We are now interested in examining the scenario where a \textit{sparse} number of point queries are corrupted with Gaussian noise. To handle this, we change the sampling scheme to random subset selection so that the $i^{th}$ measurement takes the form: $y_i = \frac{f(\vecxi_j + \epsilon\vecphi_{i,j}) - f(\vecxi_j)}{\epsilon}$. This particular formulation allows us to analyse the impact of corruption of a sparse number of queries with Gaussian noise, along the directions specified by $\vecphi$. We use the sparCS algorithm with non convex constraints \cite{sparCS} for the recovery of the low rank matrix $X$ (defined in Section \ref{subsec:lowrank_recov}). We choose the parameters $d,m_{\setX}$ and $k$ identically as in the previous experiment. Additionally we choose the sparsity parameter to be $1\%$ of the number of measurements $m_{\Phi}$, i.e. for each value of $m_{\Phi}$, $1\%$ of the measurements are corrupted with Gaussian noise. The standard deviation of the noise, $\sigma$ is set to 0.01 as previously. By varying $d$ from 30 to 120 in steps of 15, we compute the minimum number of measurements $m_{\Phi}$ needed to achieve $\frac{1}{k}\norm{\matA\widehat{\matA}}^2_F \ \geq \ 0.95$. We observe that for each $d$, we require to sample around 90\% of the entries of the matrix $X$ to achieve the desired approximation performance. Figure \ref{fig:ridge_quad_sparse_noise} shows that $m_{\Phi}$ scales approximately linearly with the dimension $d$.

%
%
\section{Conclusions}\label{sec: conc}
In this work, we consider the problem of learning multi-ridge functions of the
form $f(\vecx) = g(\matA\vecx)$, for arbitrary $\matA \in \Real^{k \times d}$
where rank($\matA$) = $k$. As compared to ~\cite{Fornasier2010} we make no
compressibility assumption on the rows of $\matA$ thus generalizing their work
to arbitrary $\matA$. Assuming $g$ to be a $\mathcal{C}^2$ function, our
learning strategy leverages a generic stable low rank matrix recovery program to
first recover an approximation $\widehat{\matA}$ to $\matA$ (up to an
orthonormal transformation), and then uses $\widehat{\matA}$ to form an
approximation to $f$. We emphasize that our theoretical learning guarantees are
algorithm independent as long as the low rank recovery algorithm is stable. We
then establish the sampling complexity of our approach to be polynomial in the
dimension $d$. We also provide local conditions that enable us to capture basis
free sparse additive models within our framework.

Interesting future directions would involve sampling schemes for
$\mathcal{C}^{r}$ functions with $0 < r < 2$, thus removing the current
requirement that the ridge function to be approximated belong to the
$\mathcal{C}^2$ class. Moreover, studying the minimax sampling lowerbounds for
our approximation problem is also important. Finally, we hope to tie our
analysis with the regression setting.
\section*{Acknowledgements}
This work was supported in part by the European Commission under Grant MIRG-268398, 
ERC Future Proof,  SNF 200021-132548, SNF 200021-146750 and SNF CRSII2-147633.
VC also would like to acknowledge Rice
University for his Faculty Fellowship. The authors thank Jan Vybiral for useful
discussions and Anastasios Kyrillidis for helping with simulations.
\bibliographystyle{plain}
\bibliography{ridge_functions_references}

\appendix
\section{Low-rank recovery formulations} \label{sec: ap: lowrank}
We consider three distinct low-rank recovery problem settings, depending on $\Phi$ and $E$:

\paragraph{\bf 1.\ Affine rank minimization (ARM):} The ARM problem is exactly \eqref{factorization_equation_gen}, where $\Phi$ is a general linear operator and the $i$-th entry of $\vecy$ is obtained via $[\Phi(\matX)]_i \ = \ll \Phi_i, \matX\gg$. Over the last decade, several convex and non-convex algorithms address the perturbed ARM problem, such as nuclear norm minimization, matrix Dantzig selector, singular value thresholding, and ADMIRA  \cite{RechtFazel2010,Raghu2009,CandesRecht2009,CandesTao2010,lee2010admira}

\paragraph{\bf 2.\ Matrix completion (MC):} The MC problem revolves around a modification of \eqref{factorization_equation_gen} as follows
\begin{equation}\label{eq: MC}
\vecy = \Phi_{\Omega}\left(\matX + E(\setX,\epsilon,\setPhi_{\Omega})\right),
\end{equation}
where $\Phi_{\Omega}$ is a subset selection operator that samples a set of entries $\matX_{i,j}, (i,j)\in \Omega$, ($\abs{\Omega}=m_{\Phi}$) within the complete set of entries $[d]\times [m_{\setX}]$. The ARM algorithms also handle the MC problem.

\paragraph{\bf 3.\ Robust principal component analysis (RPCA):} The original RPCA problem assumes that $\Phi$ is the identity operator so that we observe all the entries of $\matX$. Recent generalizations also address the ARM and MC sampling formulations. An important difference compared to ARM and MC models, however, is that the RPCA approach explicitly handles unbounded outliers in observations (i.e., $\vecs(\pi)$ in \eqref{eq: noisy oracle}).\footnote{Here, we constrain the RPCA formulation to only the case where $\Phi$ is a subset selection operator as in \eqref{eq: MC}.} We highlight two RPCA algorithms, which relies on a convex formulation \cite{candes2009robust}, and sparCS, which explicitly carries non-convex rank and sparsity constraints \cite{sparCS}.

%

%
\section{Proof of Proposition ~\ref{proposition:noise_bounded_proof_gen}} \label{sec:bounded_noise_proof}
\begin{proof}
By definition:
\begin{align*}
\norm{\vecnoise}^2_{l_2^{m_{\Phi}}} &= \frac{\epsilon^2}{4}\left(\sum_{i=1}{m_{\Phi}}\abs{\sum_{j=1}^{m_{\setX}}\vecphij^T \nabla^2f(\veczeta_{i,j})\vecphij}^2 \right).
\end{align*}
Then, the following holds true:
\begin{align*}
\abs{\vecphij^T \nabla^2f(\veczeta_{i,j})\vecphij} &= \abs{\vecphij^T A^T \nabla^2 g(\matA\veczeta_{i,j})\matA\vecphij} \nonumber \\ &\leq \norm{\nabla^2 g(\matA\veczeta_{i,j})}_F \norm{\matA \vecphij}^2_{l^k_2} \leq \frac{k^2 C_2 d}{m_{\Phi}}.
\end{align*}
Therefore,
\begin{equation*}
\norm{\vecnoise}^2_{l_2^{m_{\Phi}}} \leq \frac{\epsilon^2}{4}\left(\sum_{i=1}^{m_{\Phi}}\left( \frac{m_{\setX}k^2 C_2 d}{m_{\Phi}}\right)^2\right)
= \frac{\epsilon^2}{4}\frac{m_{\setX}^2k^4C_2^{2}d^2}{m_{\Phi}}.
\end{equation*}
\end{proof}
%
\section{Proof of Lemma \ref{lemma:adjoint_noise_bound}} \label{sec:specnorm_adj_noise_proof}
\begin{proof}
Let $E = \Phi^{*}(\vecnoise)$. We have
$\norm{\Phi^{*}(\vecnoise)} = \text{sup}_{v,w \in \mathbb{S}^{m_{\setX}-1}}\abs{\left\langle v,Ew\right\rangle}.$
\begin{eqnarray*}
\left\langle v,Ew\right\rangle &= & \text{Tr}(v^TEw) \ = \ \text{Tr}(Ewv^T) \\
&=& \text{Tr}(\Phi^{*}(\vecnoise)wv^T) \ = \ \left\langle vw^T,\Phi^{*}(\vecnoise)\right\rangle \ \\
&=& \left\langle\Phi(vw^T) ,\vecnoise\right\rangle \leq  \norm{\vecnoise}_{l_2^{m_{\Phi}}}\norm{\Phi(vw^T)}_{l_2^{m_{\Phi}}}.
\end{eqnarray*}
Using Proposition ~\ref{proposition:noise_bounded_proof_gen} and since $\norm{\Phi(vw^T)}_{l_2^{m_{\Phi}}}^2 \leq (1+\kappa_1)$ holds with probability at 
least $1-2e^{-m_{\Phi}q(\kappa_1)+(d+m_{\setX}+1)u(\kappa_1)}$, we arrive at the stated bound on $\norm{\Phi^{*}(\vecnoise)}$.
\end{proof}
\section{Proof of Corollary  \ref{corollary:rank_k_approximate}}\label{sec: ap: cor4}
\begin{proof}
Lemma ~\ref{lemma:adjoint_noise_bound} in conjunction with Theorem ~\ref{theorem:Dantzig_recovery} gives us the following bound on $\norm{\matX-\widehat{\matX}_{DS}}_{F}^2$:
\begin{equation}
\norm{\matX-\widehat{\matX}_{DS}}_{F}^2 \leq \frac{C_0 C_2^2 k^5 \epsilon^2 d^2 m_{\setX}^2}{4m_{\Phi}}(1+\kappa). \label{eq:rank_k_recovery}
\end{equation}
In general, we can have $\text{rank}(\widehat{\matX}_{DS}) > k$, thus we consider the best rank $k$ approximation to $\widehat{\matX}_{DS}$, in the sense of $\norm{\cdot}_F$. 
We then obtain the following error bound:
\begin{eqnarray*}
\norm{\matX-\widehat{\matX}^{(k)}_{DS}}_F \leq \norm{\matX-\widehat{\matX}_{DS}}_F + \norm{\widehat{\matX}_{DS}-\widehat{\matX}^{(k)}_{DS}}_F
&\leq 2\norm{\matX-\widehat{\matX}_{DS}}_F.
\end{eqnarray*}
Here, $\norm{\widehat{\matX}_{DS}-\widehat{\matX}^{(k)}_{DS}}_F \leq \norm{\matX-\widehat{\matX}_{DS}}_F$ as $\widehat{\matX}^{(k)}_{DS}$ is the 
best rank $k$ approximation to $\widehat{\matX}_{DS}$ in the sense of $\norm{\cdot}_F$. Finally using ~\eqref{eq:rank_k_recovery} we arrive at the stated bound.
\end{proof}

\section{Proof of Lemma \ref{lemma:mata_approximation}} \label{sec:proof_lemma_mata_approx}
Before beginning the proof of Lemma \ref{lemma:mata_approximation} we first recall the following theorem by ~\cite{Tropp2010}, 
which provides bounds on the deviation behaviour of the largest and smallest eigenvalues of the sum of independent positive semidefinite random matrices.
\begin{prop}(Matrix Chernoff) \label{prop:tropp_chernoff_bound}
Consider $\matX_1,\dots,\matX_m$ independent positive semidefinite random matrices of dimensions $k \times k$. 
Assume that $\lambda_1(X_j) \leq C$, where $\lambda_1(X_j) \geq \dots \geq \lambda_k(X_j)$ represent the eigenvalues of $X_j$. Denote the eigenvalues of the sum of the expectations as
\begin{equation*}
\lambda_{\max} = \lambda_1\left(\sum_{j=1}^{m}\mathbb{E}[X_j]\right) \quad \text{and} \quad \lambda_{\min} = \lambda_{k}\left(\sum_{j=1}^{m}\mathbb{E}[X_j]\right).
\end{equation*}
Then, we have the following so-called user-friendly bounds
\begin{equation*}
\mathbb{P}\left(\left\{\lambda_k\left(\sum_{j=1}^{m}X_j\right) \leq (1-\rho)\lambda_{\min}\right\}\right) \leq k\exp\left(-\frac{\lambda_{\min}\rho^2}{2C} \right), \forall \rho \in (0,1),
\end{equation*}
\begin{equation*}
\mathbb{P}\left(\left\{\lambda_1\left(\sum_{j=1}^{m}X_j\right) \geq (1+\rho)\lambda_{\max} \right\}\right) \leq k\left(\frac{1+\rho}{e}\right)^{\frac{-\lambda_{\max}(1+\rho)}{C}}, \forall \rho \in ((e-1),\infty).
\end{equation*}
\end{prop}
We now provide the proof of Lemma \ref{lemma:mata_approximation} below.
%
\begin{proof}
Observe that by Weyls inequality ~\cite{Weyl1912} we have $\abs{\widehat{\sigma}_l - \sigma_l} < \tau$. Assuming $\tau < \sigma_k$ we have
\begin{equation*}
\min_l \{\sigma_l, \widehat{\sigma}_l\} \geq (\sigma_k - \tau).
\end{equation*}
Thus by applying Wedins perturbation bound ~\cite{Wedin1972} we obtain the following bound on $\norm{\matA^T\matA - \widehat{\matA}^T\widehat{\matA}}_F$:
\begin{align*}
\norm{\matA_1^T\matA_1 - \widehat{\matA}^T\widehat{\matA}}_F = \norm{\matA^T\matA - \widehat{\matA}^T\widehat{\matA}}_F \leq \frac{2}{(\sigma_k - \tau)}\norm{\matX - \widehat{\matX}^{(k)}_{DS}}_F \leq \frac{2\tau}{(\sigma_k - \tau)}.
\end{align*}
We also have the following simplified expression for $\norm{\matA^T\matA - \widehat{\matA}^T\widehat{\matA}}_F$:
\begin{align*}
\norm{\matA^T\matA - \widehat{\matA}^T\widehat{\matA}}_F^2 = 2k - 2\text{Tr}(\matA^T\matA\widehat{\matA}^T\widehat{\matA}) = 2k - 2\norm{\matA\widehat{\matA}^T}_F^2.
\end{align*}
This leads to the following lower bound on $\norm{\matA\widehat{\matA}^T}_F$:
\begin{align}
2k - 2\norm{\matA\widehat{\matA}^T}_F^2 \leq \frac{4\tau^2}{(\sigma_k - \tau)^2}
\Leftrightarrow \norm{\matA\widehat{\matA}^T}_F \geq \left(k - \frac{2\tau^2}{(\sigma_k - \tau)^2} \right)^{1/2}. \label{eq:A_mat_approx}
\end{align}
For a non-trivial bound on $\norm{\matA\widehat{\matA}^T}_F$, we require the following to hold true:
\begin{align}
k - \frac{2\tau^2}{(\sigma_k - \tau)^2} > 0
\Leftrightarrow \sqrt{\frac{k}{2}} > \frac{\tau}{(\sigma_k - \tau)}
\Leftrightarrow \tau < \frac{\sigma_k\sqrt{\frac{\displaystyle k}{\displaystyle 2}}}{(1+\sqrt{\frac{\displaystyle k}{\displaystyle 2}})}. \label{eq:tau_cond}
\end{align}
Applying Proposition ~\ref{prop:tropp_chernoff_bound} on ~\eqref{eq:matrix_sum} and observing that $C = k C_2^2$, 
we have with probability at least $1-k\exp\left(-\frac{\displaystyle m_{\setX}\alpha \rho^2}{\displaystyle 2k C_2^2} \right)$ that 
$\lambda_{k}\left(\sum_{j=1}^{m} X_j \right) \geq (1-\rho) m_{\setX}\alpha$ or equivalently $\sigma_k \geq \sqrt{(1-\rho)m_{\setX}\alpha}$ holds true. 
Thus conditioning on the above event, we see that \eqref{eq:tau_cond} is ensured if
\begin{equation}
\tau < \left(\frac{\sqrt{(1-\rho)m_{\setX}\alpha k}}{\sqrt{k}+\sqrt{2}} \right). \label{eq:tau_final_cond}
\end{equation}
Also, plugging the above bound on $\sigma_k$ in ~\eqref{eq:A_mat_approx} we obtain the stated bound on $\norm{\matA\widehat{\matA}^T}_F$. Lastly, observe that ~\eqref{eq:tau_final_cond} is ensured if
\begin{equation*}
\epsilon < \frac{\displaystyle 1}{\displaystyle C_2 k^2 d (\sqrt{k}+\sqrt{2})}\left(\frac{\displaystyle (1-\rho) m_{\Phi}\alpha}{\displaystyle (1+\kappa) C_0 m_{\setX}}\right)^{1/2}.
\end{equation*}
\end{proof}
%
\section{Proof of Theorem \ref{thm:main_approx_thm}}\label{sec: ap: proof of main theorem}
\begin{proof}
We first observe that: $\widehat{f}(\vecx) \ = \ f(\widehat{\matA}^T\widehat{\matA}\vecx) \ = \ g(\matA\widehat{\matA}^T\widehat{\matA}\vecx)$.
\begin{align*}
\therefore \abs{f(\vecx) - \widehat{f}(\vecx)} = \abs{g(\matA\vecx) - g(\matA\widehat{\matA}^T\widehat{\matA}\vecx)}
&\leq C_2\sqrt{k}\norm{(\matA - \matA\widehat{\matA}^T\widehat{\matA})\vecx}_{l_2^k} \\
&\leq C_2\sqrt{k}\norm{\matA - \matA\widehat{\matA}^T\widehat{\matA}}_F \norm{\vecx}_{l_2^d}.
\end{align*}
Now it is easy to verify that:
\begin{align*}
\norm{\matA - \matA\widehat{\matA}^T\widehat{\matA}}_F^2 = \text{Tr}((\matA^T - \widehat{\matA}^T\widehat{\matA}\matA^T)(\matA - \matA\widehat{\matA}^T\widehat{\matA})) = k - \norm{\matA\widehat{\matA}^T}_F^2.
\end{align*}
Using Lemma ~\ref{lemma:mata_approximation} and the fact that $\norm{\vecx}_{\ell_2^d} \leq 1+\bareps$, we arrive at the stated approximation bound. 
Finally, to establish the claim in terms of $\delta$ in Theorem \ref{thm:main_approx_thm}, we work our way backwards from the approximation guarantee and obtain the stated bounds.
\end{proof} 
\section{The relation of $\alpha$ to the hessian of $f$}{\label{subsec:tractability_gen}}
In our Ansatz, we define $\alpha$ to be a lower bound on the smallest singular value of $H^f$ in \eqref{eq:mat_cond_param}. 
Therefore, $\alpha$ is also the smallest singular value of the following matrix:
\begin{equation*}
H^{g} := \int_{\mathbb{S}^{d-1}}\nabla g(\matA\vecx) \nabla g(\matA\vecx)^{T} d\mu_{\mathbb{S}^{d-1}}(\vecx).
\end{equation*}
We now note that the uniform measure $\mu_{\mathbb{S}^{d-1}}$ on the sphere $\mathbb{S}^{d-1}$ is a rotation invariant measure. 
For instance, if we were to project the standard rotation invariant Gaussian measure on $\mathbb{R}^d$  onto $\mathbb{S}^{d-1}$ through: 
$\vecx \mapsto \vecx/\norm{\vecx} \ ; \vecx \in \mathbb{R}^d / \set{\veco}$, then the resulting measure would also be rotation invariant, 
whereby coinciding with $\mu_{\mathbb{S}^{d-1}}$. We also observe that if we were to project the measure $\mu_{\mathbb{S}^{d-1}}$ through any 
$k \times d$ matrix $\matA$ with orthonormal rows then the resultant measure $\mu_k$ is also rotation invariant and does not depend on the choice of $\matA$.

It is a well known fact that the push-forward measure of $\mu_{\mathbb{S}^{d-1}}$ on the unit ball $B_{\mathbb{R}^k}$ is given by
\begin{equation*}
\mu_{k} = \frac{\Gamma(\frac{d}{2})}{\pi^{k/2}\Gamma(\frac{d-k}{2})}(1-\norm{\vecy}_{l_2^k}^2)^{\frac{d-k-2}{2}}\mathcal{L}^{k}.
\end{equation*}
A proof of the above can be found for example in Section 1.4.4 of ~\cite{Rudin1980} where the case $\mathbb{C}^n$ is considered, which also covers the 
case $\mathbb{R}^n$. Based on this argument, we now arrive at the following equivalent expression for $H^{g}$:
\begin{equation*}
H^{g} := \frac{\Gamma(\frac{d}{2})}{\pi^{k/2}\Gamma(\frac{d-k}{2})}\int_{B_{\mathbb{R}^k}}\nabla g(\vecy) \nabla g(\vecy)^T (1-\norm{\vecy}_{l_2^k}^2)^{\frac{d-k-2}{2}} d\vecy.
\end{equation*}

If the dimension $d \rightarrow \infty$ and if $k$ is fixed, the measure $\mu_k$ concentrates around 0 exponentially fast. That is, for an open ball $B_{\Real_k}(\epsilon)$ 
for a fixed $\epsilon\in(0,1)$, we have
\begin{equation*}
\mu_k(B_{\Real_k}(\epsilon)) \rightarrow 1, \quad \text{exponentially fast as} \quad d \rightarrow \infty.
\end{equation*}
This phenomenon is the classical concentration of the measure $\mu_{\mathbb{S}^{d-1}}$ for large dimension $d$. Informally stated, the measure $\mu_{\mathbb{S}^{d-1}}$ 
concentrates around the equator of $\mathbb{S}^{d-1}$ as $d \rightarrow \infty$. This in turn results in the concentration of the measure $\mu_k$ around a ball of smaller 
and smaller radius in $\mathbb{R}^k$. We can therefore intuitively observe that the conditioning of the matrix $H^g$ for large $d$ would be determined predominantly by the 
behavior of $g$ in a open neighborhood around the origin.
\begin{remark}
If the function $f$ is of the form $f(\vecx) = g(\matA\vecx + \vecb)$ then the expression for $H^g$ becomes the following
\begin{equation*}
H^{g} := \frac{\Gamma(\frac{d}{2})}{\pi^{k/2}\Gamma(\frac{d-k}{2})}\int_{B_{\mathbb{R}^k}}\nabla g(\vecy+\vecb) \nabla g(\vecy+\vecb)^T (1-\norm{\vecy}_{l_2^k}^2)^{\frac{d-k-2}{2}} d\vecy .
\end{equation*}
Denoting $B_{\Real_k}(\vecb,\epsilon)$ to be an open neighborhood around $\vecb$ for some $0 < \epsilon < 1$, we see that $\mu_k(B_{\Real_k}(\vecb,\epsilon)) \rightarrow 1$ as 
$d \rightarrow \infty$. In other words, the conditioning of the matrix $H^g$ would now depend on the smoothness properties of $g$ in an open neighborhood of the point $\vecb$. 
Keeping this in mind, we can take $\vecb$ to be $\veco$ without loss of generality.
\end{remark}
\section{Proof of Proposition \ref{prop:alpha_tract_gen}}\label{sec: ap: pprop2}
%
\begin{proof}
Denote $\frac{\displaystyle \partial g}{\displaystyle \partial y_i} = \gideriv$ and $\frac{\displaystyle \partial^2 g}{\displaystyle \partial y_i\partial y_j} = \gijderiv$. 
By writing the Taylor's series of $\gideriv$ and $\gjderiv$ around $\veco$ we obtain
\begin{align*}
\gideriv(\vecy) &= \gideriv(\veco) + \sum_{l=1}^{k} y_l \gilderiv(\veczeta_i), \\
\gjderiv(\vecy) &= \gjderiv(\veco) + \sum_{l=1}^{k} y_l \gjlderiv(\veczeta_j)
\end{align*}
where $\veczeta_i,\veczeta_j$ depend on $\vecy$. 
Denote $H^g_{i,j}$ as the $(i,j)^{th}$ entry of $H^g$. We now obtain the following expression for $H^g_{i,j}$:
\begin{equation}\label{eq: Hessian}
H^g_{i,j} = h_1+h_2+h_3,
\end{equation}
where
\begin{equation}\label{eq: grad term}
  h_1= \gideriv(\veco)\gjderiv(\veco),
\end{equation}
\begin{align}
  h_2 = \frac{\Gamma(\frac{d}{2})}{\pi^{k/2}\Gamma(\frac{d-k}{2})}[\gideriv(\veco)&\sum_{l_2=1}^{k}\int_{B_{\Real^k}}y_{l_2}\gjltderiv(\veczeta_j)(1-\norm{\vecy}_{\ell_2^k}^2)^{\frac{d-k-2}{2}}d\vecy + \nonumber \\
  &\gjderiv(\veco)\sum_{l_1=1}^{k}\int_{B_{\Real^k}}y_{l_1}\giloderiv(\veczeta_i)(1-\norm{\vecy}_{\ell_2^k}^2)^{\frac{d-k-2}{2}}d\vecy],  
\end{align}
and
\begin{equation}
  h_3 = \frac{\Gamma(\frac{d}{2})}{\pi^{k/2}\Gamma(\frac{d-k}{2})}\sum_{l_1,l_2=1}^{k}\int_{B_{\Real^k}}y_{l_1}y_{l_2}\giloderiv(\veczeta_i)\gjltderiv(\veczeta_j)(1-\norm{\vecy}_{\ell_2^k}^2)^{\frac{d-k-2}{2}}d\vecy.
\end{equation}
%
We first focus on the term $h_3$. For some $0 < \theta < 1$, let $\mathcal{U}_{\theta} = B_{\Real^k}(\theta)$ denote an open neighborhood of the origin. 
Then due to concentration of measure phenomenon, $\mu_k(\mathcal{U}_{\theta}) \rightarrow 1$ as $d \rightarrow \infty$, typically exponentially fast. 
Hence for large $d$ we have the following approximation for $h_3$, where the approximation error decays exponentially fast with dimension (see the end of the proof for the rates):
\begin{align}
\therefore \ h_3 &\approx  \frac{\Gamma(\frac{d}{2})}{\pi^{k/2}\Gamma(\frac{d-k}{2})}\sum_{l_1,l_2=1}^{k}\int_{\mathcal{U}_{\theta}} 
y_{l_1}y_{l_2}\giloderiv(\veczeta_i)\gjltderiv(\veczeta_j)(1-\norm{\vecy}_{\ell_2^k}^2)^{\frac{d-k-2}{2}}d\vecy \nonumber \\
&= B_{d,k} \sum_{l_1,l_2=1}^{k} I_{il_1,jl_2}(d,k) \label{eq:hgij_exp}
\end{align}
where $I_{il_1,jl_2}(d,k) = \int_{\mathcal{U}_{\theta}} y_{l_1}y_{l_2}\giloderiv(\veczeta_i)\gjltderiv(\veczeta_j)(1-\norm{\vecy}_{\ell_2^k}^2)^{\frac{d-k-2}{2}}d\vecy$ 
and $B_{d,k} = \frac{\Gamma(\frac{d}{2})}{\pi^{k/2}\Gamma(\frac{d-k}{2})}$. 
Now from the Lipschitz continuity of $\frac{\displaystyle \partial^2 g}{\displaystyle \partial y_i \partial y_j}(\vecy)$ in $\mathcal{U}_{\theta}$ we have:
\begin{equation}
\abs{\gijderiv(\vecy) - \gijderiv(\veco)} < \theta L ; \quad i,j = 1,\dots k, \quad \forall \vecy \in \mathcal{U}_{\theta} \label{eq:lip_cond}
\end{equation}
Using \eqref{eq:lip_cond} it is easy to verify the following for $\veczeta_i, \veczeta_j \in \mathcal{U}_{\theta}$:
\begin{equation}
\giloderiv(\veco)\gjltderiv(\veco) - C \le \giloderiv(\veczeta_i)\gjltderiv(\veczeta_j) \le \giloderiv(\veco)\gjltderiv(\veco) + C, \label{eq:gijderiv_bounds}
\end{equation}
where $C = L^2\theta^2 + 2C_2\theta L$. We now proceed to upper bound $h_3$ by first considering $I_{il_1,jl_2}(d,k)$:

\begin{align*}
I_{il_1,jl_2}(d,k) = \int_{\mathcal{U}_{\theta}:y_{l_1} y_{l_2} > 0} &y_{l_1}y_{l_2}\giloderiv(\veczeta_i)\gjltderiv(\veczeta_j)(1-\norm{\vecy}_{\ell_2^k}^2)^{\frac{d-k-2}{2}}d\vecy \ + \\
&\int_{\mathcal{U}_{\theta}:y_{l_1} y_{l_2} < 0} y_{l_1}y_{l_2}\giloderiv(\veczeta_i)\gjltderiv(\veczeta_j)(1-\norm{\vecy}_{\ell_2^k}^2)^{\frac{d-k-2}{2}}d\vecy
\end{align*}
Using ~\eqref{eq:gijderiv_bounds} we arrive at the following upper bound:
\begin{align*}
I_{il_1,jl_2}(d,k) \le \int_{\mathcal{U}_{\theta}} &y_{l_1}y_{l_2}\giloderiv(\veco)\gjltderiv(\veco)(1-\norm{\vecy}_{\ell_2^k}^2)^{\frac{d-k-2}{2}}d\vecy \ + \\
&2C \int_{\mathcal{U}_{\theta}:y_{l_1}y_{l_2} > 0} y_{l_1}y_{l_2}(1-\norm{\vecy}_{\ell_2^k}^2)^{\frac{d-k-2}{2}}d\vecy
\end{align*}
Plugging the above bound on $I_{il_1,jl_2}(d,k)$ in ~\eqref{eq:hgij_exp} we get:

\begin{align}
h_3 &\lesssim B_{d,k} \sum_{l_1,l_2=1}^{k} \int_{\mathcal{U}_{\theta}} y_{l_1}y_{l_2}\giloderiv(\veco)\gjltderiv(\veco)(1-\norm{\vecy}_{\ell_2^k}^2)^{\frac{d-k-2}{2}}d\vecy \ + \nonumber \\
&2C B_{d,k} \sum_{l_1,l_2=1}^{k}\int_{\mathcal{U}_{\theta}:y_{l_1}y_{l_2} > 0} y_{l_1}y_{l_2}(1-\norm{\vecy}_{\ell_2^k}^2)^{\frac{d-k-2}{2}}d\vecy \nonumber \\
&\le B_{d,k} \sum_{l=1}^{k} \gilderiv(\veco)\gjlderiv(\veco) \int_{\mathcal{U}_{\theta}}
y^2_{l}(1-\norm{\vecy}_{\ell_2^k}^2)^{\frac{d-k-2}{2}}d\vecy \ + \nonumber \\
&2C B_{d,k} \sum_{l_1,l_2=1}^{k}\int_{\mathcal{U}_{\theta}} (y_{l_1}^2 + y_{l_2}^2)(1-\norm{\vecy}_{\ell_2^k}^2)^{\frac{d-k-2}{2}}d\vecy \quad \left((y_{l_1}^2 + y_{l_2}^2)/2 \geq y_{l_1}y_{l_2}\right) \nonumber \\
&=  \left(\frac{1}{k}\sum_{l=1}^{k}\gilderiv(\veco)\gjlderiv(\veco) + 4Ck \right) B_{d,k}\int_{\mathcal{U}_{\theta}}
\norm{\vecy}^2(1-\norm{\vecy}_{\ell_2^k}^2)^{\frac{d-k-2}{2}}d\vecy \label{eq:hgij_upper}
\end{align}
Proceeding similarly one can obtain the following lower bound:
\begin{equation}
h_3 \gtrsim \left(\frac{1}{k}\sum_{l=1}^{k}\gilderiv(\veco)\gjlderiv(\veco) - 4Ck \right) B_{d,k}\int_{\mathcal{U}_{\theta}}
\norm{\vecy}^2(1-\norm{\vecy}_{\ell_2^k}^2)^{\frac{d-k-2}{2}}d\vecy. \label{eq:hgij_lower}
\end{equation}
%
We now focus on the term $h_2$. Similar to before, we have the following approximation for $h_2$, where the approximation error decays exponentially fast with dimension.
\begin{align} \label{eq:h2_approx}
  h_2 \approx \frac{\Gamma(\frac{d}{2})}{\pi^{k/2}\Gamma(\frac{d-k}{2})}[\gideriv(\veco)&\sum_{l_2=1}^{k}\int_{\mathcal{U}_{\theta}}y_{l_2}\gjltderiv(\veczeta_j)(1-\norm{\vecy}_{\ell_2^k}^2)^{\frac{d-k-2}{2}}d\vecy + \nonumber \\
  &\gjderiv(\veco)\sum_{l_1=1}^{k}\int_{\mathcal{U}_{\theta}}y_{l_1}\giloderiv(\veczeta_i)(1-\norm{\vecy}_{\ell_2^k}^2)^{\frac{d-k-2}{2}}d\vecy],  
\end{align}
Now it is easily verifiable that
\begin{align}
\int_{\mathcal{U}_{\theta}}y_{l_2}\gjltderiv(\veczeta_j)(1-\norm{\vecy}_{\ell_2^k}^2)^{\frac{d-k-2}{2}}d\vecy &= 
\int_{\mathcal{U}_{\theta}:y_{l_2}>0}y_{l_2}\gjltderiv(\veczeta_j)(1-\norm{\vecy}_{\ell_2^k}^2)^{\frac{d-k-2}{2}}d\vecy + \nonumber \\
&\int_{\mathcal{U}_{\theta}:y_{l_2}<0}y_{l_2}\gjltderiv(\veczeta_j)(1-\norm{\vecy}_{\ell_2^k}^2)^{\frac{d-k-2}{2}}d\vecy \nonumber \\
&< 2\theta L \int_{\mathcal{U}_{\theta}:y_{l_2}>0} y_{l_2} (1 - \norm{\vecy}_{\ell_2^k}^2)^{\frac{d-k-2}{2}}d\vecy \label{eq:h2_term1_up_bound}
\end{align}
where \eqref{eq:h2_term1_up_bound} follows by making use of \eqref{eq:lip_cond}. Through a similar process on the second summation term 
in \eqref{eq:h2_approx} and by using $\abs{\gideriv(\veco)}, \abs{\gjderiv(\veco)} < C_2$ one obtains the following upper bound on $h_2$.
\begin{equation} \label{eq:h2_upper_bound}
h_2 \lesssim \frac{\Gamma(\frac{d}{2})}{\pi^{k/2}\Gamma(\frac{d-k}{2})}[4k C_2\theta L \int_{\mathcal{U}_{\theta}} \norm{\vecy}(1-\norm{\vecy}_{\ell_2^k}^2)^{\frac{d-k-2}{2}}d\vecy]
\end{equation}
One can similarly verify the following lower bound on $h_2$.
\begin{equation} \label{eq:h2_lower_bound}
h_2 \gtrsim \frac{-\Gamma(\frac{d}{2})}{\pi^{k/2}\Gamma(\frac{d-k}{2})}[4k C_2\theta L \int_{\mathcal{U}_{\theta}} \norm{\vecy}(1-\norm{\vecy}_{\ell_2^k}^2)^{\frac{d-k-2}{2}}d\vecy].
\end{equation}
Lastly the integral term in the above bound can be bounded from above as follows.
%
\begin{align*}
\int_{\mathcal{U}_{\theta}} \norm{\vecy}(1-\norm{\vecy}_{\ell_2^k}^2)^{\frac{d-k-2}{2}}d\vecy &= \frac{2\pi^{k/2}}{\Gamma(\frac{k}{2})} \int_{0}^{\theta} r^k (1-r^2)^{(d-k-2)/2} dr \\
&< \frac{2\pi^{k/2}}{\Gamma(\frac{k}{2})} \int_{0}^{1} r^{k-1} (1-r^2)^{(d-k-2)/2} dr \\
&= \frac{\pi^{k/2} \Gamma(\frac{d-k}{2})}{\Gamma(\frac{d}{2})}.
\end{align*}
Using this in \eqref{eq:h2_upper_bound} and \eqref{eq:h2_lower_bound} we obtain:
\begin{equation} \label{eq:h2_upper_lower_bound}
-4k C_2\theta L \lesssim h_2 \lesssim 4k C_2\theta L.
\end{equation}
%
By re-writing ~\eqref{eq:hgij_upper}, ~\eqref{eq:hgij_lower}, \eqref{eq:h2_upper_lower_bound} and combining with \eqref{eq: grad term}  we obtain \eqref{eq: Hessian} in matrix form:
\begin{align}\label{eq: Hessian final}
H^g &\precsim \nabla g(\veco) \nabla g(\veco)^{T}+ 4k C_2\theta L\mathbf{1}\mathbf{1}^T + 4Ck C_{d,k}\mathbf{1}\mathbf{1}^T + \frac{C_{d,k}}{k} \nabla^2 g(\veco)\nabla^2 g(\veco)^T, \\
H^g &\succsim \nabla g(\veco) \nabla g(\veco)^{T}- 4k C_2\theta L\mathbf{1}\mathbf{1}^T - 4Ck C_{d,k}\mathbf{1}\mathbf{1}^T+ \frac{C_{d,k}}{k} \nabla^2 g(\veco)\nabla^2 g(\veco)^T\\
\end{align}
where $C_{d,k} \ := \  B_{d,k}\int_{\mathcal{U}_{\theta}}
\norm{\vecy}^2(1-\norm{\vecy}_{\ell_2^k}^2)^{\frac{d-k-2}{2}}d\vecy$ and $\mathbf{1}$ is a $k \times 1$ vector of all ones.

Now, we show that $C_{d,k} = \Theta(1/d)$ as $d \rightarrow \infty$. By the change of variables: $r = \norm{\vecy}$, we obtain
\begin{equation*}
C_{d,k} = \frac{2\Gamma\left(\frac{d}{2}\right)}{\Gamma\left(k/2\right)\Gamma\left(\frac{d-k}{2}\right)}\int_{0}^{\theta} r^{k+1} (1-r^2)^{\frac{d-k-2}{2}}dr.
\end{equation*}
It can be checked that:
\begin{equation}
\int_{0}^{\theta}r^{k+1}(1-r^2)^{\frac{d-k-2}{2}}dr \leq \int_{0}^{1}r^{k+1}(1-r^2)^{\frac{d-k-2}{2}}dr = \frac{1}{2}\left[\frac{\Gamma(\frac{d-k}{2})\Gamma(\frac{k+2}{2})}{\Gamma(\frac{d+2}{2})}\right]. \label{eq:upper_bd}
\end{equation}
One can also verify that:
\begin{align}
\int_{\theta}^{1}r^{k+1}(1-r^2)^{\frac{d-k-2}{2}}dr \leq \int_{\theta}^{1}r^{k-1}(1-r^2)^{\frac{d-k-2}{2}}dr \leq e^{-\left(\frac{d-k-2}{2}\right)\theta^2} \nonumber \\
\Rightarrow \int_{0}^{\theta}r^{k+1}(1-r^2)^{\frac{d-k-2}{2}}dr \geq 
\frac{1}{2}\left[\frac{\Gamma(\frac{d-k}{2})\Gamma(\frac{k+2}{2})}{\Gamma(\frac{d+2}{2})}\right] - e^{-\left(\frac{d-k-2}{2}\right)\theta^2}. \label{eq:lower_bd}
\end{align}
From \eqref{eq:upper_bd} and \eqref{eq:lower_bd} we get the following bounds for $C_{d,k}$:
\begin{align*}
C_{d,k} \le \frac{k}{d}, \quad C_{d,k} \ge \left(\frac{k}{d} -\frac{2\Gamma\left(\frac{d}{2}\right)}{\Gamma\left(k/2\right)\Gamma\left(\frac{d-k}{2}\right)} e^{-\left(\frac{d-k-2}{2}\right)\theta^2} \right).
\end{align*}
In other words, $C_{d,k} = \Theta(k/d)$ as $d \rightarrow \infty$, 
for fixed $k, \theta$.\footnote{$\theta$ can depend on $k$, which is not a problem since $k$ is fixed.}

In \eqref{eq: Hessian final}, we have a summation of four terms. The first three terms are rank-$1$ matrices with the last two vanishing as $d$ grows. 
The fourth term is a full rank matrix by assumption. In this case, denote $\matV$ as the summation of $\matD= \frac{C_{d,k}}{k} \nabla^2 g(\veco)\nabla^2 g(\veco)^T$  
and the rank-$1$ matrix $\matE = \nabla g(\veco) \nabla g(\veco)^{T}$: $\matV= \matD+\matE$. Since both matrices are symmetric positive semidefinite, 
we can use the singular value interlacing theorem for rank-$1$ perturbations \cite{WSo1999}, which states
\begin{equation}\label{eq: interlacing}
  \sigma_1(\matV) \ge \sigma_1(\matD) \ge \sigma_2(\matV) \ge \sigma_2(\matD) \ge \ldots \ge \sigma_{k-1}(\matD) \ge \sigma_k(\matV) \ge \sigma_k(\matD).
\end{equation}
Therefore, the order of the $k$-th largest singular value of $\matV$ is bounded by the $(k-1)$-th and the $k$-th largest singular values of $\matD$, 
which scale as $C_{d,k}$. In other words, $\sigma_k(\matV) = \Theta(1/d)$.

Moreover, using results for eigenvalue bounds for symmetric interval matrices \cite{Rohn2005}, we have the following bounds on the singular values of $H^g$:
\begin{equation} \label{eq:final_eig_bounds_Hg}
\sigma_i(\matV) - 4Ck^2 C_{d,k} - 4C_2\theta L k^{2} \leq \lambda_{i}(H^g) \leq \sigma_i(\matV) + 4Ck^2 C_{d,k} + 4C_2\theta L k^{2}.
\end{equation}
where we recall that $C = L^2\theta^2 + 2C_2\theta L$. We now consider the following scenarios:

\begin{enumerate}
\item If $\nabla g(\veco) = \veco$, then the ``$4C_2\theta L k^{2}$ term'' in \eqref{eq:final_eig_bounds_Hg} vanishes, leading to  
\begin{equation*}
 \lambda_{k}(H^g) \in \left[\sigma_k(\matV) - \frac{4Ck^3}{d} , \sigma_k(\matV) + \frac{4Ck^3}{d}\right].
\end{equation*}
Hence for $\theta = O(1/k^3)$, we obtain $\lambda_{i}(H^g) = \Theta(1/d)$.

\item If $\nabla g(\veco) \neq \veco$, we obtain
\begin{equation} \label{eq:eig_bds_nz_grad}
 \lambda_{k}(H^g) \in \left[\sigma_k(\matV) - 4\left(\frac{Ck^3}{d} + C_2\theta L k^{2}\right) , \sigma_k(\matV) + 4\left(\frac{Ck^3}{d} + C_2\theta L k^{2}\right)\right].
\end{equation}
We see from \eqref{eq:eig_bds_nz_grad} that $\lambda_{k}(H^g) = \Theta(1/d)$ holds provided the Lipschitz constant $L$ is sufficiently small. 
In particular, if $L = O(1/d)$, then for $\theta = O(1/k^3)$ we see that $\lambda_{k}(H^g) = \Theta(1/d)$ holds true.
\end{enumerate}

\end{proof}

\end{document}